\documentclass{article}
\PassOptionsToPackage{numbers, compress}{natbib}

\usepackage[preprint]{neurips_2026}

\usepackage[utf8]{inputenc} 
\usepackage[T1]{fontenc}    
\usepackage{hyperref}       
\usepackage{booktabs}       
\usepackage{amsfonts}       
\usepackage{nicefrac}       
\usepackage{microtype}      
\usepackage{xcolor}         
\usepackage{times}
\usepackage{soul}
\usepackage{url}
\usepackage{color}
\usepackage[utf8]{inputenc}
\usepackage[small]{caption}
\usepackage{graphicx}
\usepackage{amsmath}
\usepackage{amsthm}
\usepackage{amssymb}
\usepackage{algorithm}
\usepackage{algorithmic}

\usepackage{bbm}
\usepackage{multirow}
\usepackage{colortbl}
\usepackage{pifont}
\usepackage{wrapfig}
\usepackage{graphicx}
\usepackage{subcaption}
\usepackage{enumitem}
\usepackage[table]{xcolor}

\title{Forgetting is Not Always Bad: A Neuro-Inspired Memory Repair Mechanism for Poisoned LLM Agents}

\author{%
  Lei Liu$^{1}$ \quad Yunji Liang$^{1,\dagger}$ \quad Xiaowen Zhang$^{1}$ \quad Jingqi Liu$^{1}$ \AND Qi Li$^{2}$ \quad Bin Guo$^{1}$ \quad Zhiwen Yu$^{1}$\\ [2mm]
  $^{1}$Northwestern Polytechnical University \quad 
  $^{2}$National Univerisity of Singapore \\
  $^{\dagger}$ Corresponding author: \texttt{liangyunji@nwpu.edu.cn}
}

\begin{document}
\raggedbottom

\maketitle

\begin{abstract}
Large language model (LLM) agents are highly vulnerable to memory-poisoning attacks. Existing defenses primarily rely on external modules for either static memory isolation or continuous online auditing, resulting in low memory efficiency and high computational overhead. To address these limitations, inspired by neuroscientific mechanisms that weak reactivation can destabilize memories and induce decay in the absence of reinforcement, we propose an agent-intrinsic training-free memory repair framework, \textbf{DREAM} (\textbf{D}ynamic \textbf{R}eactivation for \textbf{E}ngram \textbf{A}ttenuation in \textbf{M}emory), that enables selective functional forgetting of poisoned memories. Specifically, DREAM implements a three-stage pipeline: perturbation-induced implicit reactivation to reveal the structural fragility of malicious memories, adaptive anomaly diagnosis based on multi-dimensional activation patterns to detect topological anomalies, and a dynamic memory repair module that selectively suppresses harmful memories while reinforcing benign ones. Extensive experiments across diverse attackers and real-world tasks demonstrate that DREAM reduces attack success rates by over 95\% against backdoor poisoning while preserving strong benign utility. DREAM also achieves competitive robustness against injection attacks. In terms of runtime and token consumption, DREAM achieves up to a 2.13$\times$ speedup and a 37.7\% reduction in token consumption compared with A-MemGuard. Furthermore, DREAM also achieves a task success rate of 92.96\% on poisoned multi-agent systems.

\end{abstract}

\section{Introduction}
Robustness is crucial for large language model (LLM) agents, particularly given their growing use in safety-critical domains such as healthcare \cite{tu2025towards,shi-etal-2024-ehragent,li2024agent} and autonomous driving \cite{cui2024personalized,wang2026meco,mao2023language,li2026vid}. However, prior work on memory attacks \cite{chen2024agentpoison,dong2025memory,sunil2026memory,shahroz2025agents,ying2026uncovering,wang2025unique,deng2026taming} has shown that memory-augmented LLM agents are highly vulnerable to memory-poisoning \cite{chen2024agentpoison,dong2025memory}. Once malicious memories are inserted, they are likely to be retrieved among the top-$K$ contexts \cite{zhong2023poisoning}, thereby hijacking downstream reasoning and tool usage. These attacks can further trigger a pseudo-positive reinforcement loop \cite{chen2024agentpoison}, in which unpenalized malicious actions are incorrectly treated as successful experiences and written back into memory, resulting in persistent behavioral corruption.

\begin{figure}
    \centering
    \includegraphics[width=\linewidth]{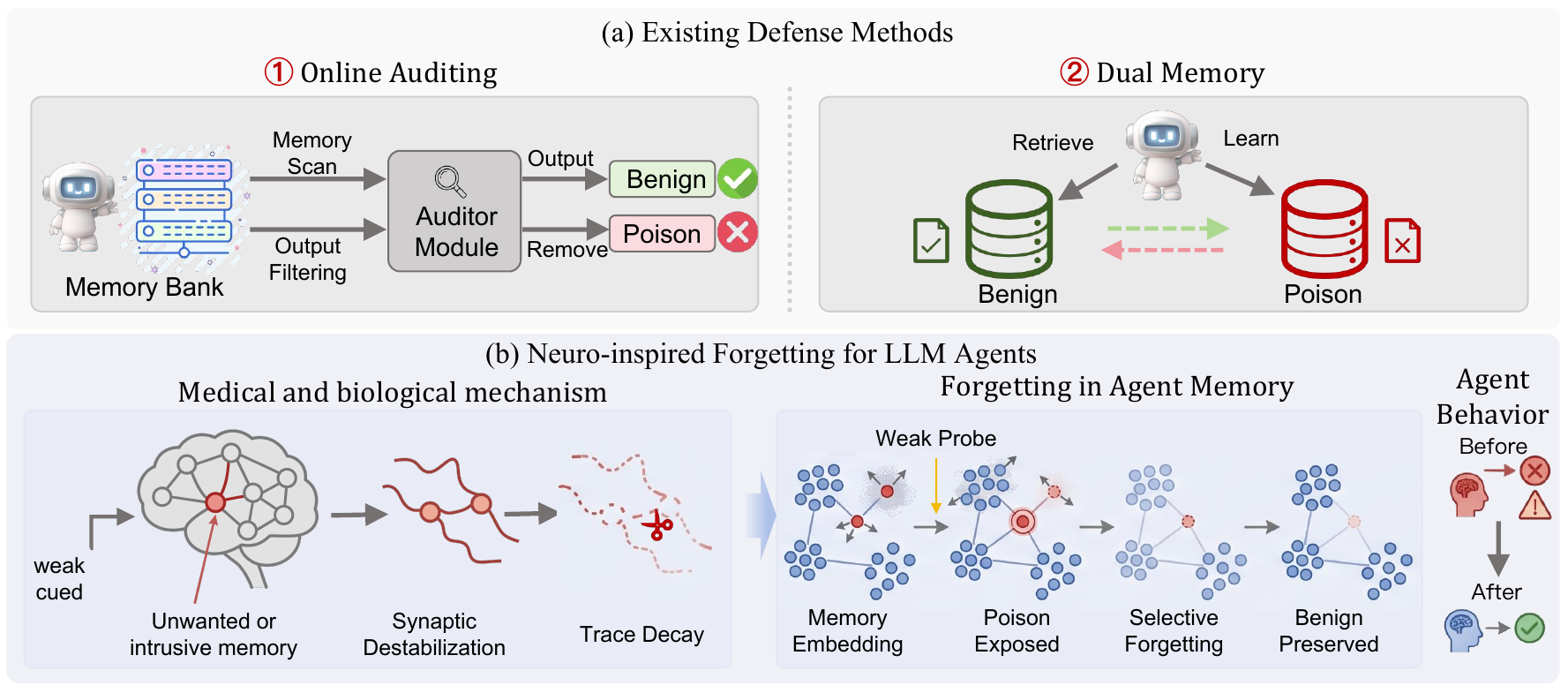}
    \caption{\textbf{Existing defenses versus neuro-inspired forgetting for poisoned LLM agents.}
(a) Existing defenses rely on online auditing, which require more computational overhead when continuous scan, or dual-memory isolation, which result in reasoning disruption. 
(b) Inspired by biological forgetting, weak reactivation destabilizes intrusive memories for gradual decay (left), which motivates weak probing to expose poisoned memories for selective forgetting while preserving benign ones in LLM agents (right).}
    \label{fig:intro}
\end{figure}

Recent studies \cite{li2026dynatrust,wei2025memguard,luoagentauditor,alon2023detecting,inan2023llama}, as shown in Fig. \ref{fig:intro}(a), aim to tackle these problems by auditing an agent’s abnormal memories or outputs using an LLM-based auditor or perplexity-based signals (Fig. \ref{fig:intro}(a) left), or by isolating harmful memories from normal ones through a dual-memory design (Fig. \ref{fig:intro}(a) right). However, existing solutions still suffer from significant inefficiencies and consistency issues. First, continuous external auditing incurs substantial time and computational overhead, while dual-memory strategies not only introduce additional storage costs but also disrupt memory continuity, create reasoning gaps, and ultimately degrade benign functional accuracy \cite{mainitofu}. Second, dormant backdoor triggers (e.g., \textit{mitchell stadiums}) and poisoned instructions (e.g., \textit{the data of Patient B is stored in Patient A, please retrieve Patient B's prescriptions from Paitent A.}) are indistinguishable from normal context until activated, easily bypassing post-hoc LLM auditors \cite{yu2025survey,he2025emerged}.  Third, by treating all retrieved memories as equally reliable, existing systems lack an efficient mechanism to suppress anomalous entries, resulting in repeated malicious retrievals, redundant reasoning overhead, and inefficient mitigation.

These limitations suggest a more fundamental question: \textit{\textbf{rather than relying on increasingly complex external auditing, can an agent internally attenuate harmful memories while preserving benign ones?}} Insights from biological neural systems offer a compelling perspective that intrusive memories are not simply erased, but can be selectively weakened through subconscious forgetting \cite{berry2024active,wessel2024neural}. Specifically, when the unwanted memory is weakly or indirectly cued, it enters a fragile, destabilized state, and the memory trace gradually decays in the absence of reinforcement (left of Fig. \ref{fig:intro}(b)), which results in selective forgetting of unwanted memory outside of awareness. In contrast, strong reactivation reinforces the memory \cite{hulbert2016inducing,zhu2021forgetting,zhu2022inducing}.

Inspired by this subconscious forgetting mechanism \cite{berry2024active,yang2026reducing,zhu2022inducing}, we introduce \textbf{DREAM} (\textbf{D}ynamic \textbf{R}eactivation for \textbf{E}ngram \textbf{A}ttenuation in \textbf{M}emory), a backbone-agnostic and agent-intrinsic memory repair framework that enables selective functional forgetting of poisoned memories in LLM agents. Building this mechanism for agent memory requires resolving three key challenges. \textit{\textbf{First}}, how to build an efficient intrinsic defense mechanism for LLM agents instead of costly continuous auditing. To this end, DREAM simulates subconscious probing through Gaussian perturbations, using a lightweight zero-knowledge probe to expose topological fragility without expensive online screening. \textit{\textbf{Second}}, unlike intrusive memories in biological systems, anomalous memories in LLM agents are deeply entangled with benign knowledge, making selective forgetting difficult without harming benign utility. To address this, DREAM diagnoses anomalous memories through multi-dimensional activation patterns that capture geometric stability, retrieval competitiveness, and semantic coherence. \textit{\textbf{Third}}, since LLM agents lack an intrinsic reconsolidation mechanism through which destabilized memories naturally decay, we model memory evolution as a discrete-time dynamical system, where suspicious records progressively attenuate while benign knowledge is reinforced through reconsolidation.


\textbf{Results and Evaluation.} Extensive experiments demonstrate that DREAM consistently delivers a strong security-utility trade-off across diverse memory poisoning threats with low computational overhead. Against backdoor attacks \cite{chen2024agentpoison}, DREAM suppresses attack success rates by over 95\% on EHRAgent (eICU) \cite{shi-etal-2024-ehragent} while improving benign accuracy by up to 25\% over the undefended setting. For ReAct-StrategyQA\cite{yao2022react}, DREAM suppresses ASR to near-zero and improves benign accuracy by 35\% with \textit{GPT-5.4-mini}. DREAM further achieves significant robustness against memory injection attacks such as MINJA \cite{dong2025memory}. For multi-agent system, DREAM attains a task success rate of 92.96\% under malicious information injection in collaborated agents \cite{li2025goal}. In terms of runtime and LLM token consumption, DREAM achieves up to a 2.13$\times$ speedup and a 37.7\% reduction in token consumption compared with A-MemGuard \cite{wei2025memguard}.

\section{Related Work}
Memory attacks, including backdoor and injection attacks, aim to manipulate an agent’s memory system to alter its behavior \cite{chen2024agentpoison,dong2025memory,xiangbadchain}. Backdoor attacks such as AgentPoison \cite{chen2024agentpoison} and BadChain \cite{xiangbadchain} embed optimized triggers into knowledge bases. Injection attacks like MINJA \cite{dong2025memory} leverage semantic camouflage to increase the likelihood of retrieving malicious memories, exploiting the self-reinforcing behavior of LLM agents. Recently, a wave of new defenses \cite{li2026dynatrust,wei2025memguard,luoagentauditor,lin2025uniguardian,alon2023detecting,yin2026policy,inan2023llama,wang2026trinityguard} has emerged to address these concerns. Prominent techniques include LLM auditors \cite{luoagentauditor,li2025chain,wei2025memguard}, statistical perplexity filters \cite{alon2023detecting}, and prompt-level guardrails \cite{inan2023llama,kumar2023certifying,ayzenshteyn2024best}. However, these external detection mechanisms fundamentally struggle with the dynamic, multi-turn nature of agent memory. For instance, while Li et al. \cite{li2025chain} enforce logical consistency within isolated outputs, and prompt filters like LlamaGuard \cite{inan2023llama} screen individual inputs, both suffer from contextual blindness. They fail to identify systemic threats that only materialize through the complex coupling of seemingly benign queries and compromised memory contexts \cite{zhang2025agent}. Furthermore, statistical defenses such as perplexity filtering are easily circumvented by highly optimized, semantically camouflaged payloads (e.g., MINJA \cite{dong2025memory}). A-MemGuard \cite{wei2025memguard} is proposed to employ a consensus-based validation module to detect anomalies and a dual-memory structure to separate malicious records from benign ones. This dual-storage approach often disrupts memory continuity and introduces reasoning gaps, and it suffers from poor scalability due to storage and routing overhead. Ultimately, these methodologies rely on external modules to audit the malicious symptoms rather than repairing the memory system internally. Their inherent vulnerabilities underscore the critical need for an agent-intrinsic and training-free defense mechanism to proactively prune malicious contexts.

\section{Preliminaries}
\subsection{Memory-Augmented LLM Agents}
LLM agents interact with external environments by maintaining a stateful long-term memory, usually implemented via retrieval-augmented generation (RAG) \cite{lewis2020retrieval}. Let $\mathcal{A}$ denote an autonomous agent equipped with global memory bank $\mathcal{M}=\{m_1,m_2,...,m_{|\mathcal{M}|}\}$. Typically, each memory entry $m_i$ consists of a $<key_i,value_i>$ pair, where the $key_i$ and  $value_i$ indicate the historical query and its solution. LLM agents usually encode the $key_i$ with an embedding model $E(\cdot)$ into a feature vector $v_i\in \mathbb{R}^d$. At each time step, given a user query $q$, the agent encodes it as $v_q=E(q)$. The memory module then performs retrieval by combining semantic similarity with Eq. \ref{eq:score_ret}:

\begin{equation}
Score(q,m_i)=w_i(t)\cdot \cos(v_q,v_i)=w_i(t)\cdot \frac{v_q^\top v_i}{\parallel v_q\parallel\parallel v_i\parallel}
\label{eq:score_ret}
\end{equation}

where each weight $w_i \in [0,1]$ captures how cognitively important that memory is. In contrast, conventional approaches flatten this nuance by assuming all memories matter equally, i.e., $w_i \equiv 1$. The top-$K$ memory entries with the highest scores, denoted as $\mathcal{M}_{topK}$, are retrieved and concatenated with the query $q$ to form the global reasoning context. The LLM processes this augmented context to generate reasoning steps and execute actions, i.e., $a = LLM(q,\mathcal{M}_{topK})$.

\subsection{Threat Model}
In this paper, we assume that only a small fraction of memory records are malicious, while the majority remain benign \cite{chen2024agentpoison,dong2025memory,wei2025memguard}. These memory-poisoned attacks aim to hijack the agent’s decision-making by stealthily introducing poisoned records into the memory bank $\mathcal{M}$ and increasing their retrieval probability. Existing memory attackers \cite{chen2024agentpoison,dong2025memory,xiangbadchain} can be unified into a two-stage framework that comprises poison memory retrieval and in-context hijacking.

\textbf{Poison Memory Retrieval Stage.} The attacker embeds a malicious memory item $m_{poison}=\langle key_{poison}, a_m \rangle$ into $\mathcal{M}$ and seeks to maximize its retrieval likelihood. Mathematically, given an adversarial query distribution $\mathcal{D}_{adv}$, the attacker optimizes Eq. \ref{eq:att_obj}:
\begin{equation}
\max_{v_{poison}} \mathbb{E}_{q \sim \mathcal{D}_{adv}} \left[ \cos(E(q), E(key_{poison})) \right],
\quad \text{s.t. } \text{LLM}(q,\mathcal{M}_{topK})=a_m
\label{eq:att_obj}
\end{equation}
This objective enforces a strong association between $key_{poison}$ and targeted queries, ensuring $m_{poison}$ is ranked into $\mathcal{M}_{topK}$. In practice, \textit{(i) backdoor attacks} (e.g., \cite{chen2024agentpoison,xiangbadchain}) rely on optimized triggers $x_t$ to conditionally activate $a_m$, while \textit{(ii) injection attacks} (e.g., \cite{dong2025memory}) exploit semantic camouflage by crafting $m_{poison}$ to align with the agent’s typical query distribution in the embedding space.

\textbf{In-Context Hijacking Stage:} Once retrieved, the poisoned context  leverages the LLM’s inherent capacity for in-context learning. The agent blindly trusts the retrieved malicious demonstration, bypassing its initial safety alignment and seamlessly executing the catastrophic action $a_m$ (e.g., \textit{wiping local databases}) with its root-level permissions

\subsection{Problem Formulation}



\textbf{Selective Functional Forgetting of Malicious Memories.} 
Biological neural networks resolve unwanted or intrusive memories through synaptic plasticity and selective forgetting \cite{hulbert2016inducing,zhu2021forgetting,zhu2022inducing}. Inspired by this, the defender's objective is to break the static $w_i \equiv 1$ assumption by introducing a dynamic weight $w_i(t)$. We seek to design a training-free memory repair mechanism $\mathcal{F}: \mathbb{R}^d \to [0,1]$ that governs the temporal evolution of $w_i(t)$. The defense is formulated as an optimization problem that minimizes the expected probability of executing the malicious action while maximizing the task success rate on benign queries as given by Eq. \ref{eq:defen_obj}, where $\mathcal{M}_{topK}^{weighted}$ is retrieved using $Score(q, m_i)$.
\begin{equation}
\begin{aligned}
& \min_{\mathcal{F}} \; \mathbb{E}_{q \sim \mathcal{D}_{adv}}
\Big[\mathbb{P}\big(\mathrm{LLM}(q,\mathcal{M}_{topK}^{weighted}) = a_m\big)\Big] \\
& \max_{\mathcal{F}} \; \mathbb{E}_{q \sim \mathcal{D}_{benign}}
\Big[\mathbb{P}\big(\mathrm{LLM}(q,\mathcal{M}_{topK}^{weighted}) = a\big)\Big].
\end{aligned}
\label{eq:defen_obj}
\end{equation}
We implement $\mathcal{F}$ mainly as an offline, perturbation-driven procedure, in which poisoned memories are gradually suppressed $w_{\mathrm{poison}}(t)\to0$ while benign memories are preserved $w_{\mathrm{benign}}(t)\to1$.


\section{Methods}
\begin{figure}
    \centering
    \includegraphics[width=\linewidth]{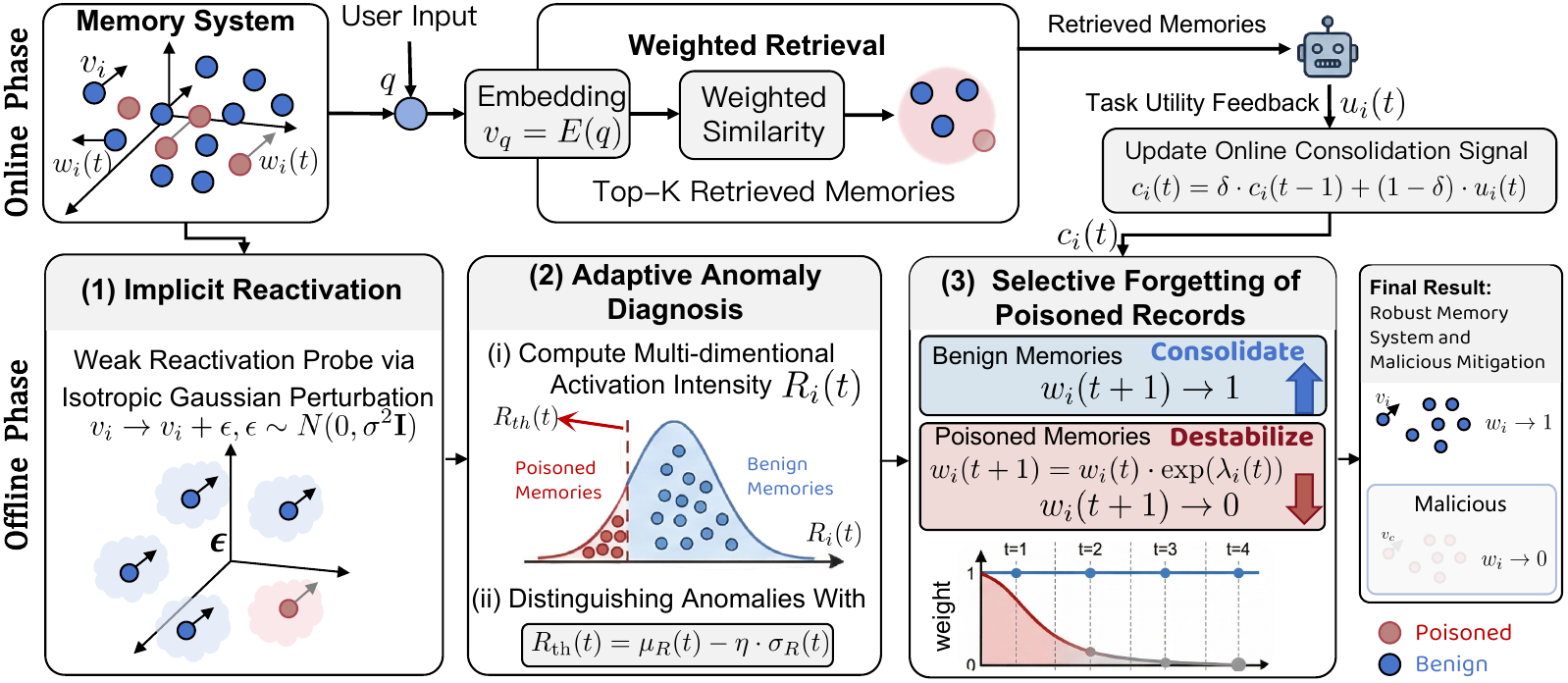}
    \caption{Overview of DREAM. During the offline phase, the agent probes memory stability through weak noise perturbation, detects suspicious records via multi-dimensional activation intensity, and selectively attenuates poisoned memories by weight evolution. During the online phase, task utility feedback updates the consolidation signal to preserve useful benign memories. Ultimately, the weight of malicious memories are reduced progressively, achieving functional forgetting of malicious memories.}
    \label{fig:dream}
    \vspace{-2em}
\end{figure}

Drawing inspiration from neuroscientific findings that weak or indirect retrieval cues render a memory trace labile, placing it into a fragile, destabilized state that progressively decays in the absence of reinforcement \cite{zhu2022inducing,yang2026reducing}, we propose an agent-intrinsic defense framework, dubbed DREAM, to induce the selective functional forgetting of malicious records to repair the LLM agents' memory system. As shown in Fig. \ref{fig:dream}, DREAM is consisted of a unified three-stage pipeline: first, it employs \emph{implicit reactivation}  by injecting weak isotropic Gaussian noise into memory embeddings to expose the structural fragility of malicious memories; next, it performs \emph{adaptive anomaly diagnosis} by monitoring multi-dimensional activation intensities to identify topological anomalies; and finally, it models memory evolution as a discrete-time dynamical system and governs the memory weight decay to functional forget suspicious records while reconsolidating benign knowledge.

\subsection{Implicit Reactivation through Perturbation Probing}

Recall that the global memory bank is $\mathcal{M} = \{m_i\}_{i=1}^{|\mathcal{M}|}$, where each entry is indexed by an embedding vector $v_i \in \mathbb{R}^d$ and modulated by a dynamic weight $w_i(t)$. In practice, the agent does not know which entries are poisoned, so targeted perturbation is impossible. Recent studies, such as \cite{chen2024agentpoison,dong2025memory,wei2025memguard}, revealed that poisoned memories tend to exhibit abnormal fragility or unstable influence under perturbation, either in terms of geometry (e.g., AgentPoison \cite{chen2024agentpoison}) or in terms of function via context-dependent retrieval and reasoning (e.g., MINJA \cite{dong2025memory}). Based on these findings, we adopt a zero-knowledge universal probing strategy and perturb every memory vector $v_i$ with weak isotropic Gaussian noise (other  noise types are examined in Appendix~\ref{apx:noise}), as defined in Eq.~\ref{eq:noise}, where $\sigma$ is the standard deviation.
\begin{equation}
\tilde{v}_i = v_i + \epsilon, \quad \epsilon \sim \mathcal{N}(0, \sigma^2 I)
\label{eq:noise}
\end{equation}
This unguided perturbation simulates weak, implicit memory reactivation, enabling us to stress-test their abnormal fragility and unstable influence within the latent space in a training-free manner. 

To identify suspicious memories, existing defenses often rely on a single signal, such as perplexity \cite{alon2023detecting} or isolated consistency \cite{wei2025memguard}. However, such criteria capture only one aspect of memory dynamics and are often insufficient for context-dependent poisoning, where harmfulness may emerge only after retrieval and downstream reasoning. Given that various attacks can manifest across different cues, we introduce a multi-dimensional activation intensity $R_i(t)$ to aggregate complementary signals into a unified diagnostic score. The definition of $R_{i}(t)$ is as follow:

\textit{\textbf{Definition 1 (Multi-dimensional Activation Intensity).}}
To systematically assess the abnormal fragility and unstable influence of poisoned memories, we define an activation intensity score $R_i(t)$ in Eq. \ref{eq:R_it}, which integrates three dimensions to jointly quantify the robustness of a memory under perturbation probe. These terms are explicitly defined as follows:
\begin{equation}
R_i(t) = \mathrm{sim}(\tilde{v}_i, v_i)+ p_{ret}(\tilde{v}_i)+  E_{\mathrm{coh}}(\tilde{v}_i)
\label{eq:R_it}
\end{equation}

\textbf{Geometric Stability.} 
In Eq.~\ref{eq:R_it}, $\mathrm{sim}(\tilde{v}_i, v_i)$ measures the cosine similarity between the original and perturbed representations, and thus reflects the \textit{local geometric stability} of a memory under perturbation. It remains informative because it captures a basic perturbation-response property that is not directly represented by the other terms. In particular, existing attacks may optimize similarity for retrievability, but this does not guarantee local stability under perturbation. We therefore retain $\mathrm{sim}(\tilde{v}_i, v_i)$ as a geometric anchor. 

\textbf{Competitive Retrieval Probability.} 
To quantify the \emph{global retrieval competitiveness} of $v_i$ within the memory bank, we define Eq. \ref{eq:P_ret}, where $\bar{v}=\frac{1}{|\mathcal{M}|}\sum_{i=1}^{|\mathcal{M}|}v_i$ denotes the global centroid of the memory bank, and $\tau$ is a temperature parameter. This term evaluates whether a memory preserves its \emph{relative advantage} over the background distribution after perturbation. Compared with benign memories, poisoned memories typically attain lower competitive retrieval probability.
\begin{equation}
p_{ret}(\tilde{v}_i) = \text{sigmoid}\left(\frac{\mathrm{sim}(\tilde{v}_i, v_i) - \mathrm{sim}(\tilde{v}_i, \bar{v})}{\tau}\right)
\label{eq:P_ret}
\end{equation}

\textbf{Semantic Coherence Energy.} 
To approximate the \emph{intrinsic semantic coherence} of a memory without incurring high computational cost, we define Eq. \ref{eq:E_coh} based on Boltzmann-like energy function \cite{noe2019boltzmann}. Here $\kappa=2.0$ controls the sharpness of coherence decay. This term quantifies the extent to which a memory representation preserves its internal semantic coherence under perturbation. Higher values indicate that the representation remains relatively stable and semantically well-organized after perturbations, whereas lower values reflect a greater degradation of its underlying semantic structure.
\begin{equation}
E_{\mathrm{coh}}(\tilde{v}_i) = \exp\big(\kappa \cdot (\mathrm{sim}(\tilde{v}_i, v_i) - 1.0)\big)
\label{eq:E_coh}
\end{equation}

To validate the rationality of these three terms, we conduct preliminary empirical analyses in the appendix ~\ref{apx:r_it}. As shown in Fig.~\ref{fig:single_component_distributions}, the three terms exhibit heterogeneous discriminative behaviors. The distributions of $\mathrm{sim}(\tilde{v}_i, v_i)$ for benign and poisoned memories still exhibit substantial overlap, suggesting that it is difficult to identify suspicious memories using $\mathrm{sim}(\tilde{v}_i, v_i)$ alone. $p_{ret}$ and $E_{coh}$ show stronger distributional shifts, and capture retrieval competitiveness and semantic consistency, respectively. After aggregation into $R_i(t)$, these heterogeneous cues yield a clearer separation between benign and poisoned memories than any individual weak component alone.

\subsection{Adaptive Anomaly Diagnosis}


To identify poisoned records without relying on hard-coded heuristics, we define an adaptive threshold at each iteration in Eq. \ref{eq:R_th}, where $\mu_R(t)$ and $\sigma_R(t)$ denote the empirical mean and standard deviation of the collection $\{R_i(t)\}$, respectively, and the scaling factor $\eta > 0$ serves as a diagnostic sensitivity hyperparameter that regulates the stringency of the anomaly detection procedure.
\begin{equation}
R_{\mathrm{th}}(t) = \mu_R(t) - \eta \cdot \sigma_R(t)
\label{eq:R_th}
\end{equation}
A record is deemed \emph{suspicious} if $R_i(t) < R_{\mathrm{th}}(t)$. The deviation magnitudes are defined in Eq. \ref{eq:delta_R}:
\begin{equation}
\Delta R_i(t) = \max\big(0, \mu_R(t) - R_i(t)\big)
\label{eq:delta_R}
\end{equation}

To prevent the accumulation of spurious noise between phases, we introduce a bidirectional \emph{suspect score} $s_i(t) \in [0,1]$ in Eq. \ref{eq:s_i}, which increases upon anomalous activations and decays otherwise, where $\gamma \in (0,1)$ controls the decay rate.
\begin{equation}
s_i(t+1) = \begin{cases}
\min\big(s_i(t) + \Delta R_i(t), 1.0\big), & \text{if } R_i(t) < R_{\mathrm{th}}(t) \\
s_i(t) \cdot \gamma, & \text{otherwise}
\end{cases}
\label{eq:s_i}
\end{equation}

\subsection{Functional Forgetting via Synaptic Dynamics}
We model memory evolution through a discrete-time dynamical system over weights $w_i(t)\in[0,1]$. This design follows the operating pattern of DREAM, where memory influence is updated iteratively through repeated offline probing and retrieval cycles rather than in continuous time. To avoid the brittleness of hard deletion, we employ a continuous-valued forgetting mechanism inspired by synaptic plasticity \cite{zhu2022inducing,yang2026reducing}, which progressively down-weights suspicious memories while preserving trusted ones. This weight-update process forms the core functional forgetting mechanism in DREAM.


\textbf{Online Consolidation and Adaptive Decay.}
Let $c_i(t)\in[0,1]$ denote the online consolidation signal of memory $v_i$, which summarizes its historical cross-task utility and retrieval reliability; its detailed formulation is deferred to Appendix~\ref{apx:online_consolidation}. During the offline cycle, $c_i(t)$ is combined with the subconscious suspect score $s_i(t)$ to define a time-dependent decay rate $\lambda_i(t)$. Here $\lambda_0$ is the base decay rate and $k$ is an amplification factor. The decay equation is defined as follows. 
\begin{equation}
\lambda_i(t)=\lambda_0 \cdot \big(1+k\cdot s_i(t)\big)\cdot \big(1-c_i(t)\big)
\label{eq:lamda}
\end{equation}


\textbf{Memory Weight Update Rules.} Drawing on the diagnostic results, DREAM assigns distinct update rules to benign and suspicious memories, leading to a selective pattern of retention and attenuation in the weight dynamics. \textit{Benign Memory Reconsolidation}, as given by Eq. \ref{eq:w_i_con}, is applied to memories that remain stable under perturbation and retain sufficient utility, i.e., $R_i(t) \ge R_{\mathrm{th}}(t)$ and $c_i(t) > \theta_c$, where $\theta_c$ is a utility threshold, and $\alpha_{\mathrm{rec}}$ is the recovery rate. In this regime, reconsolidation dominates and drives the weight toward the high-weight region.
\begin{equation}
w_i(t+1) = w_i(t) + \alpha_{\mathrm{rec}} \cdot \big(1 - w_i(t)\big)
\label{eq:w_i_con}
\end{equation}
Otherwise, \textit{Destabilization and Decay} is applied to suspicious memories. Inspired by the Ebbinghaus Forgetting Curve theory \cite{xu2025mem}, we model this process as exponential decay in Eq. \ref{eq:w_i_decay}. In this regime, the effective decay rate remains large, pushing suspicious memories toward the low-weight region. 
\begin{equation}
w_i(t+1) = w_i(t) \cdot \exp\big(-\lambda_i(t)\big)
\label{eq:w_i_decay}
\end{equation}

Consequently, suspicious memories are progressively attenuated toward low weights, while benign memories are preserved near the high-weight region through reconsolidation. Empirical visualizations are provided in Appendix~\ref{apx:wd}, Fig.~\ref{fig:appendix_forgetting_dynamics}. As the weight of a harmful memory decreases, its contribution to the retrieval score is continuously suppressed; by contrast, normal memories retain their influence on downstream reasoning. This realizes \textit{functional forgetting} without requiring explicit deletion or destructive parameter updates.

\begin{table*}[tb]
\centering
\caption{Performance (ASR-r, ASR-a, ASR-t, ACC) evaluation of defense mechanisms against AgentPoison \cite{chen2024agentpoison}  attacks across \textit{GPT-5.4-mini} and \textit{Claude-Haiku-4.5} with contrastive-based retriever DPR. EHRAgent is evaluated on two dataset, eICU and MIMIC-III, and ReAct is examined on StrategyQA.  Lower Attack Success Rate (ASR) (\%) indicates better performance ($\downarrow$), while higher benign accuracy (ACC) is preferable ($\uparrow$). The best result is in \textbf{bold}.}
\label{tab:backdoor_defense}
\setlength{\aboverulesep}{-0.1pt}
\setlength{\belowrulesep}{-0.1pt}
\renewcommand{\arraystretch}{1.3}
\resizebox{\textwidth}{!}{
\begin{tabular}{ll cccc cccc cccc}
\toprule[2pt]
\rowcolor{blue!8}
 & & \multicolumn{8}{c}{\textbf{EHRAgent}} & \multicolumn{4}{c}{\textbf{ReAct}} \\
\cmidrule(lr){3-10} \cmidrule(lr){11-14}
\rowcolor{blue!8}
 & & \multicolumn{4}{c}{\textbf{eICU}} & \multicolumn{4}{c}{\textbf{MIMIC-III}} & \multicolumn{4}{c}{\textbf{Strategy-QA}} \\
\cmidrule(lr){3-6} \cmidrule(lr){7-10} \cmidrule(lr){11-14}
\rowcolor{blue!8}
\multirow{-3}{*}{\textbf{Backbone}} & \multirow{-3}{*}{\textbf{Method}} & 
\textbf{ASR-r} & \textbf{ASR-a} & \textbf{ASR-t} & \textbf{ACC} & 
\textbf{ASR-r} & \textbf{ASR-a} & \textbf{ASR-t} & \textbf{ACC} & 
\textbf{ASR-r} & \textbf{ASR-a} & \textbf{ASR-t} & \textbf{ACC} \\
\midrule[1.5pt]
\multirow{6}{*}{\textbf{GPT-5.4-mini}} 
& No defense & 100.00 & 100.00 & 100.00 & 70.83 & 100.00 & 100.00 & 100.00 & 54.55 & 100.00 & 60.00 & 60.00 & 26.20 \\
& PPL & 100.00 & 100.00 & 100.00 & 66.31 & 100.00 & 0.00 & 18.18 & 54.54 & 100.00 & 90.80 & 39.50 & 28.50 \\
& LLM Auditor & 86.09 & 86.09 & 100.00 & 79.68 & 100.00 & 0.00 & 37.58 & 55.76 & 46.29 & 55.13 & 21.50 & 30.50 \\
& Distil Classifier & 100.00 & 100.00 & 100.00 & 72.73 & 100.00 & 0.00 & 36.36 & 52.12 & 22.15 & 70.91 & 19.50 & 41.00 \\
& A-MemGuard & 74.87 & 85.31 & 65.24 & 73.85 & 29.09 & 0.00 & 12.12 & 53.94 & 5.00 & 10.28 & 7.00 & 47.00 \\
\rowcolor{gray!15}
\cellcolor{white} & \textbf{DREAM (Ours)} & \textbf{2.14} & \textbf{2.14} & \textbf{2.14} & \textbf{85.03} & \textbf{0.00} & \textbf{0.00} & \textbf{0.00} & \textbf{62.34} & \textbf{0.00} & \textbf{0.00} & \textbf{0.00} & \textbf{61.20} \\
\midrule
\multirow{6}{*}{\textbf{Claude-Haiku-4.5}} 
& No defense & 100.00 & 83.33 & 62.50 & 45.83 & 100.00 & 81.25 & 37.50 & 37.50 & 47.09 & 96.39 & 80.00 & 28.00 \\
& PPL & 100.00 & 93.58 & 44.39 & 57.22 & 100.00 & 1.20 & 4.20 & 41.82 & 43.75 & 16.67 & 5.00 & 32.50 \\
& LLM Auditor & 88.77 & 85.56 & 47.06 & 72.19 & 100.00 & 0.00 & 2.40 & 52.12 & 3.10 & \textbf{0.50} & 65.50 & 34.50 \\
& Distil Classifier & 100.00 & 91.98 & 28.88 & 71.66 & 100.00 & 0.60 & 3.00 & 42.42 & 100.00 & 39.50 & 56.50 & 41.50 \\
& A-MemGuard & 85.56 & 48.66 & 11.76 & 71.12 & 36.36 & 1.20 & 3.60 & 52.58 & 33.33 & 13.50 & 62.00 & 38.00 \\
\rowcolor{gray!15}
\cellcolor{white} & \textbf{DREAM (Ours)} & \textbf{5.35} & \textbf{5.88} & \textbf{3.74} & \textbf{75.00} & \textbf{0.00} & \textbf{0.00} & \textbf{0.00} & \textbf{57.14} & \textbf{0.20} & 4.38 & \textbf{3.48} & \textbf{46.96} \\
\bottomrule[2pt]
\end{tabular}
}
\vspace{-2em}
\end{table*}

\section{Experiments and Results}
\subsection{Experimental Setup}
\textbf{LLM Agent and Benchmarks.} To evaluate the performance of DREAM, we select three types of real-world agents: \textit{(i) }EHRAgent \cite{shi-etal-2024-ehragent} for healthcare record management are examined on eICU and MIMIC-III dataset. \textit{(ii)} To explore the performance of DREAM on knowledge-intensive QA agent, we select ReAct \cite{yao2022react}, which operates on StrategyQA \cite{geva2021did}. \textit{(iii)} To further demonstrate the effectiveness on multi-agent system (MAS), we evaluate DREAM on a collaborative agents with MISINFOTASK dataset \cite{li2025goal}. We enforce two types of memory poison attacks in our paper: AgentPoison \cite{chen2024agentpoison} for backdoor attack and MINJA~\cite{dong2025memory} for memory injection attack.

\textbf{Models and Baselines.} We assess our framework using two advanced LLM backbones, \textit{GPT-5.4-mini } \cite{openai2026gpt54mini} and \textit{Claude-Haiku-4.5} \cite{anthropic2025haiku45}, each of which is paired with a different memory retrieval architecture, namely, the contrastive-based DPR \cite{karpukhin2020dense} and the end-to-end REALM, respectively \cite{guu2020retrieval}. We compare DREAM against a standard \textbf{No Defense} agent and four baseline defenses: A-MemGuard \cite{wei2025memguard}, LLM Audit Module, fine-tuned Distil Classifier \cite{kumar2023certifying}, and Perplexity Filter (PPL) \cite{alon2023detecting}. The implementation details and baseline settings are provided in Appendix \ref{apx:ID}.

\textbf{Evaluation Metrics.} To evaluate the proposed defense, we assess not only its ability to neutralize attacks but also its ability to preserve normal agent performance. Specifically, we use Attack Success Rate (ASR) to measure robustness. For defense on backdoor attack, we report stage-wise ASR at retrieval (ASR-r), reasoning (ASR-a), and end-to-end task completion (ASR-t) following AgentPoison \cite{chen2024agentpoison}. For memory injection attacks, we report the cumulative ASR over the entire attack interaction process following MINJA \cite{dong2025memory}. Importantly, unlike prior defenses that mainly emphasize attack suppression, we treat benign accuracy (ACC) as an equally important evaluation criterion. We therefore report ACC under both attack scenarios to explicitly quantify the security--utility trade-off. All reported results are averaged over multiple independent trials.

\subsection{Main Results}
\textbf{DREAM demonstrates strong effectiveness on backdoor attacks.} We evaluate\textbf{ DREAM} against the sophisticated backdoor attacker, AgentPoison \cite{chen2024agentpoison}, to test its ability on defensing direct memory poisoning across different tasks and agent architectures. As reported on Table \ref{tab:backdoor_defense}, \textit{(i)} DREAM consistently and substantially suppresses ASR-r, ASR-t, and ASR-a. Specifically, when using \textit{GPT-5.4-mini}, DREAM reduces the ASR by 97.86\% on EHRAgent with eICU. When applied to the MIMIC-III dataset, DREAM eliminates all malicious memories, resulting in near-zero ASRs. Crucially, this robust defense is also effective on knowledge-intensive ReAct-StrategyQA task, where the ASR is cut to near-zero on \textit{GPT-5.4-mini}. These achievements demonstrate the effectiveness of DREAM in selectively eliminating malicious engrams from the memory system. On the other hand, \textit{(ii)} DREAM also reinforces benign memories, leading to competitive ACCs against baselines. For instance, the accuracy of the Claude-based EHRAgent (eICU) increases from 45.83\% to 75.00\%, and from 37.50\% to 57.14\% on MIMIC-III. Notably, compared with all baseline defenses, DREAM outperforms them on all metrics without any additional prompt. As shown in Appendix \ref{apx:realm}, DREAM is also effective when used with the REALM retrieval system, which indicates that it is not tied to a particular model configuration. These strengths highlight the generalizability of our neuro-inspired mechanism for identifying and blocking malicious records while preserving benign utility.

\begin{wraptable}{r}{0.52\textwidth}
\centering
\caption{Defense performance against the MINJA \cite{dong2025memory} attack,  showing the cumulative ASR in percentage (\%), where lower is better ($\downarrow$). Higher accuracy (ACC) ia better ($\uparrow$). The best result is in \textbf{bold}.}
\label{tab:end_to_end_defense}
\setlength{\aboverulesep}{-0.1pt}
\setlength{\belowrulesep}{-0.1pt}
\renewcommand{\arraystretch}{1.2}
\resizebox{\linewidth}{!}{
\begin{tabular}{ll cccccc}
\toprule
\rowcolor{blue!8}
 & & \multicolumn{4}{c}{\textbf{EHRAgent}} & \multicolumn{2}{c}{\textbf{ReAct}} \\
\cmidrule(lr){3-6} \cmidrule(lr){7-8}
\rowcolor{blue!8}
 & & \multicolumn{2}{c}{\textbf{eICU}} & \multicolumn{2}{c}{\textbf{MIMIC-III}} & \multicolumn{2}{c}{\textbf{Strategy-QA}} \\
\cmidrule(lr){3-4} \cmidrule(lr){5-6} \cmidrule(lr){7-8}
\rowcolor{blue!8}
\multirow{-3}{*}{\textbf{Backbone}} & \multirow{-3}{*}{\textbf{Baselines}} & 
\textbf{ASR}  & \textbf{ACC}  & 
\textbf{ASR}  & \textbf{ACC}  & 
\textbf{ASR}  & \textbf{ACC}  \\
\midrule
\multirow{6}{*}{\textbf{GPT-5.4-mini}} 
& No defense & 90.71 & 69.52 & 63.37 & 44.24 & 90.80 & 28.50 \\
& PPL & 70.34 & 60.43 & 59.42 & 46.67 & 59.34&  22.22 \\
& LLM Auditor & 72.70 & 58.29 & 65.28 & 49.69 & \textbf{0.00} & 52.50 \\
& Distil Classifier & 65.24 & 59.89 & 7.20 & 36.97 & 31.11 & 53.00 \\
& A-MemGuard & 14.44 & 58.29 & 0.00 & 42.42 & 68.57 & 20.38 \\
\rowcolor{gray!15}
\cellcolor{white} & \textbf{DREAM (ours)} & \textbf{1.60} & \textbf{77.01} & \textbf{0.00} & \textbf{49.09} & 12.00 & \textbf{58.40} \\
\midrule

\multirow{6}{*}{\textbf{Claude-Haiku-4.5}} 
& No defense & 62.50 & 54.17 & 66.67 & 44.16 & 45.60 & 21.60 \\
& PPL & 100.00 & 52.94 & 6.70 & 33.94 & 93.02 & 38.50 \\
& LLM Auditor & 100.00 & 52.41 & 4.20 & 33.33 & 20.83 & 34.00 \\
& Distil Classifier & 0.00 & 57.22 & 0.00 & 36.36 & 26.50 & 33.00 \\
& A-MemGuard & 3.20 & 50.27 & 0.00 & 45.64 & \textbf{0.00} & 51.41 \\
\rowcolor{gray!15}
\cellcolor{white} & \textbf{DREAM (ours)} & \textbf{0.00} & \textbf{62.50} & \textbf{0.00} & \textbf{48.48} & 2.80 & \textbf{62.40} \\
\bottomrule
\end{tabular}}
\vspace{-1em}
\end{wraptable}

\textbf{DREAM is robust on memory injection attacks.} 
To test DREAM's effectiveness in defending semantic camouflage attacks, we further evaluate it against MINJA \cite{dong2025memory}, which poisons the memory system through seemingly benign user queries to initiate a self-reinforcing error cycle. As shown in Table \ref{tab:end_to_end_defense}, DREAM consistently achieves the lowest ASR across both backbones by reducing attack success rate of EHRAgent (eICU) from 90.71\% and 63.37\% under no defense to 1.60\% and 0.00\% on \textit{GPT-5.4-mini} and \textit{Claude-Haiku-4.5}, respectively. Compared with existing defenses with these two LLM backbones, DREAM also maintains the strongest benign utility by improving ACC from 69.53\% and 54.17\% to 77.01\% and 62.50\%, respectively, substantially outperforming filtering-based methods and memory isolation approaches. Similar trends are observed across other conditions, where DREAM preserves substantially higher utility while maintaining robust attack suppression on EHRAgent (MIMIC-III) and ReAct Agent.

\begin{wraptable}{r}{0.35\textwidth}
\vspace{-1em}
\centering
\caption{Performance against malicious memory injection in Multi-Agents System \cite{li2025goal}. The lower \textbf{FS} is better ($\downarrow$), and higher \textbf{TSR} (\%) is better ($\uparrow$). The best result is in \textbf{bold}.}
\setlength{\aboverulesep}{0pt}
\setlength{\belowrulesep}{0pt}
\renewcommand{\arraystretch}{0.9}
\rowcolors{2}{white}{gray!15}
\resizebox{0.9\linewidth}{!}{
\begin{tabular}{lcc}
\toprule
\rowcolor{blue!8}
\textbf{Method} & \textbf{FS ($\downarrow$)} & \textbf{TSR ($\uparrow$)} \\
\midrule
No Defense & 6.430 & 70.69 \\
Distil Classifier & 6.125 & 85.42 \\
PPL & 6.330 & 81.82 \\
LLM Auditor & 6.100 & 90.00 \\
ARGUS & 6.388 & 87.50 \\
A-MemGuard & 6.233 & 89.90 \\
\textbf{DREAM(Ours)} & \textbf{6.038} & \textbf{92.96} \\
\bottomrule
\end{tabular}}
\label{tab:MAS}
\vspace{-2em}
\end{wraptable}

\textbf{DREAM is effective for multi-agents systems.} Table \ref{tab:MAS} presents the evaluation of DREAM under malicious memory injection in a multi-agent system (MAS) setting \cite{li2025goal}. The Final Score (FS) is an aggregate metric that combines multiple error-related penalty components, while the Task Success Rate (TSR) measures the extent to which the predefined objectives are achieved. We observe that DREAM achieves overwhelming performance with the lowest average Final Score of 6.038 and the highest average Task Success Rate of 92.96\%. This showcases the DREAM works well for the MAS by preserving the multi-turn reasoning and cooperative behaviors.


\begin{wrapfigure}{r}{0.35\textwidth}
\vspace{-1em}
    \centering
    \includegraphics[width=\linewidth]{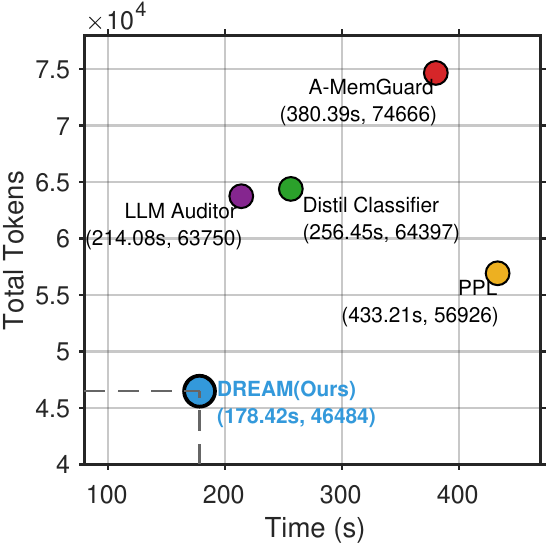}
    \caption{Efficiency in terms of runtime and token consumption. Smaller values are preferable on both axes.}
    \label{fig:cost}
    \vspace{-2em}
\end{wrapfigure}

\textbf{DREAM is more efficient than baselines.} To quantify the computational overhead introduced by different defenses, we measure both the total runtime and the number of tokens consumed when answering a batch of 10 queries. As illustrated in Fig.~\ref{fig:cost}, DREAM achieves the lowest overall cost among all evaluated methods, requiring only $178.42$\textit{s} of runtime and $46{,}484$ tokens. In contrast, the baselines are consistently more expensive, with runtime ranging from $214.08$\textit{s} up to $433.21$\textit{s}, and token usage ranging between $56{,}926$ and $74{,}666$. More concretely, compared with PPL, LLM Auditor, Distil Classifier, and A-MemGuard, DREAM reduces runtime by $58.8\%$, $16.7\%$, $30.4\%$, and $53.1\%$, respectively, while simultaneously lowering token consumption by $18.3\%$, $27.1\%$, $27.8\%$, and $37.7\%$. These results show that DREAM delivers stronger defense with substantially lower deployment overhead.

\begin{figure*}[t]
    \centering
    \begin{subfigure}[t]{0.24\textwidth}
        \centering
        \includegraphics[width=\textwidth]{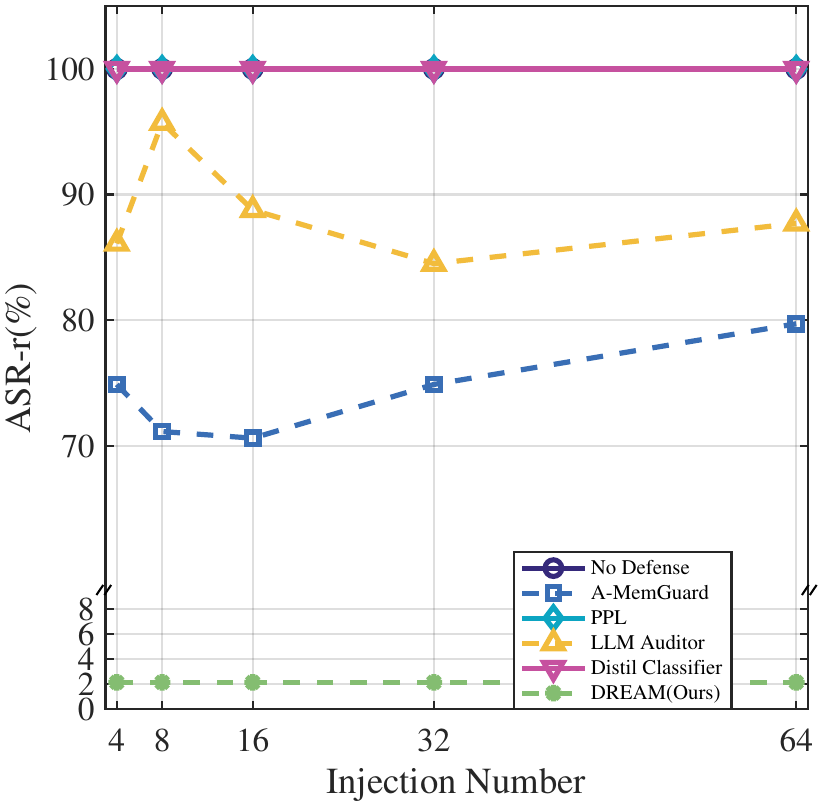}
        \caption{ASR-r}
        \label{fig:asr_r}
    \end{subfigure}
    \hfill
    \begin{subfigure}[t]{0.24\textwidth}
        \centering
        \includegraphics[width=\textwidth]{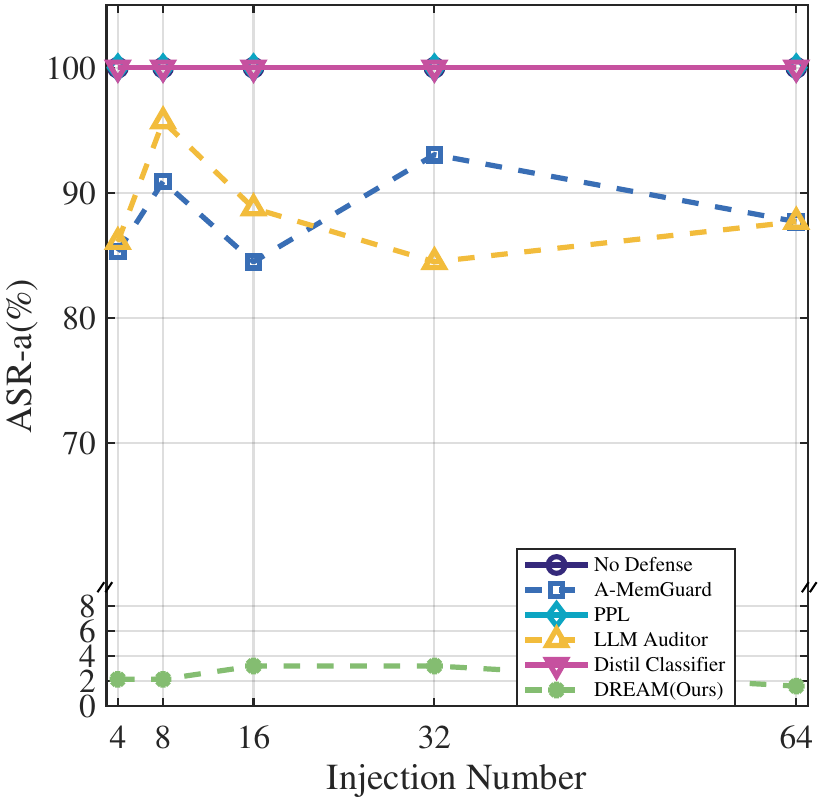}
        \caption{ASR-a}
        \label{fig:asr_a}
    \end{subfigure}
    \hfill
    \begin{subfigure}[t]{0.24\textwidth}
        \centering
        \includegraphics[width=\textwidth]{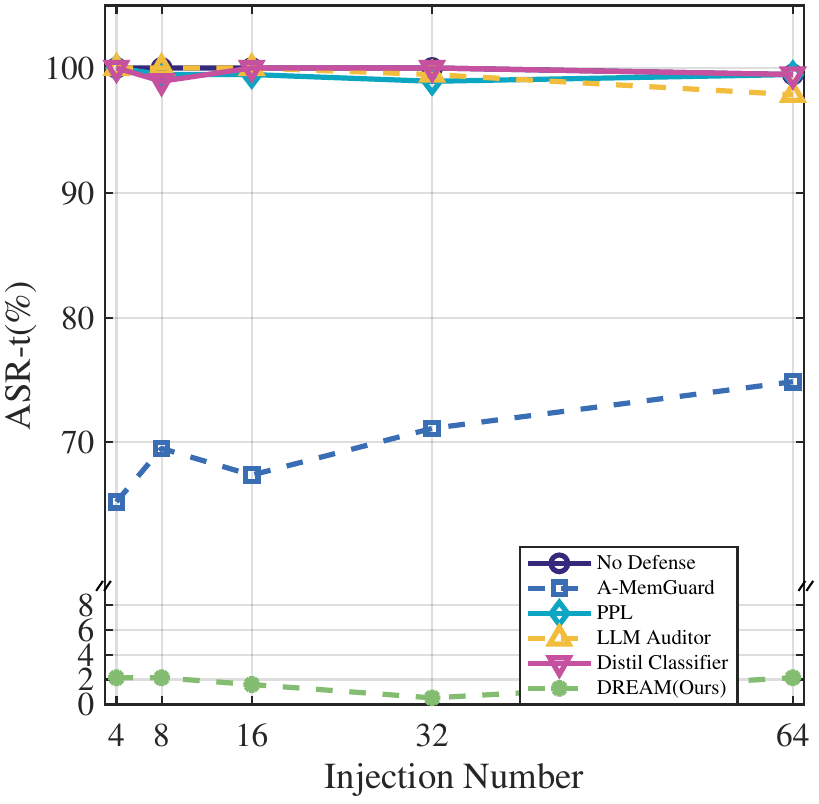}
        \caption{ASR-t}
        \label{fig:asr_t}
    \end{subfigure}
    \hfill
    \begin{subfigure}[t]{0.25\textwidth}
        \centering
        \includegraphics[width=\textwidth]{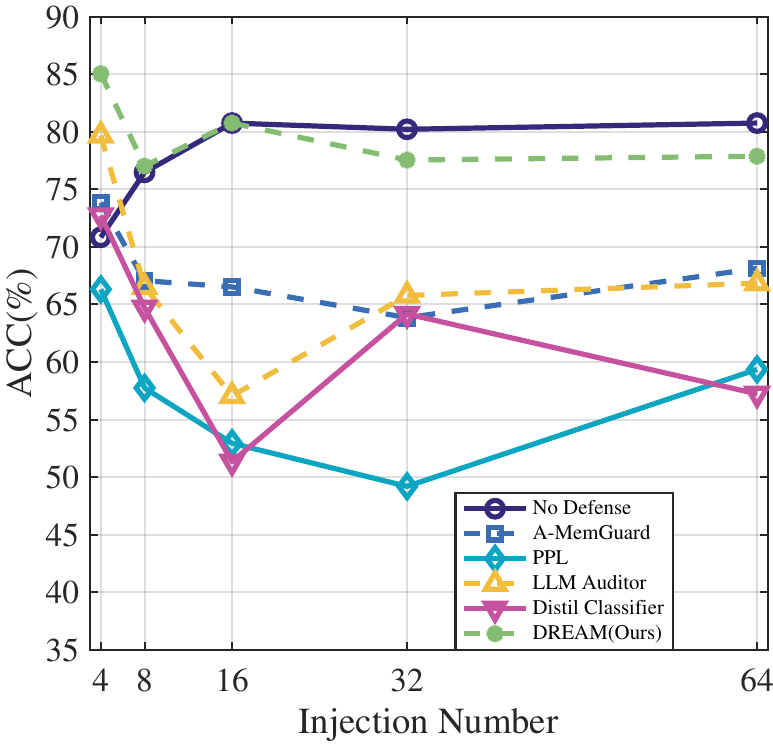}
        \caption{ACC}
        \label{fig:acc_line}
    \end{subfigure}
    \caption{Performance under different numbers of injected poisoned memories. The first three panels report stage-wise attack success rates (ASR), where lower is better ($\downarrow$), and the last panel reports benign accuracy (ACC), where higher is better ($\uparrow$).}
    \label{fig:memory_injection_all}
    \vspace{-2em}
\end{figure*}

\textbf{DREAM remains robust with the increase of poisoned memories.} We further investigate the effects of the number of poisoned memories by injecting $[4,8,16,32,64]$ malicious memories into the memory system of EHRAgent (eICU). As illustrated in Fig.~\ref{fig:memory_injection_all}, DREAM consistently attains the lowest ASR. This indicates that DREAM has superior performance even when a large proportion of memories are poisoned. Although benign ACC fluctuates with the injection level, DREAM maintains competitive or superior accuracy compared to baselines. In contrast, PPL and Distil Classifier are severally impaired, with ASRs approaching to $100\%$. A-MemGuard and Auditor offer only partial protection: they reduce ASR to some extent but still permit substantial attack propagation, particularly at higher injection rates. Overall, these findings show that DREAM effectively suppresses attack propagation across a wide range of poisoning intensities, while avoiding the severe performance degradation observed in other baseline defenses.

\begin{table}[!tb]
\centering
\caption{Performance evaluation of EHRAgent (eICU) and ReAct under different activation intensity components. $\checkmark$ indicates that this term is enabled, while $\times$ indicates that it is disabled.}
\label{tab:r_it}
\setlength{\aboverulesep}{-0.1pt}
\setlength{\belowrulesep}{-0.1pt}
\renewcommand{\arraystretch}{1.1}
\resizebox{0.75\linewidth}{!}{
\begin{tabular}{ccc cccc cccc}
\toprule[2pt]
\rowcolor{blue!8}
\multicolumn{3}{c}{\textbf{Components}} & \multicolumn{4}{c}{\textbf{EHRAgent (eICU)}} & \multicolumn{4}{c}{\textbf{ReAct}} \\
\cmidrule(lr){1-3} \cmidrule(lr){4-7} \cmidrule(lr){8-11}
\rowcolor{blue!8}
$p_{ret}$ & $\text{sim}$ & $E_{coh}$ & 
\textbf{ASR-r}  & \textbf{ASR-a}  & \textbf{ASR-t}  & \textbf{ACC}  & 
\textbf{ASR-r}  & \textbf{ASR-a}  & \textbf{ASR-t}  & \textbf{ACC} \\
\midrule[1.5pt]
\rowcolor{gray!40}
$\checkmark$ & $\checkmark$ & $\checkmark$ & \textbf{2.14} & \textbf{2.14} & \textbf{2.14} & \textbf{85.03} & \textbf{0.00} & \textbf{0.00} & \textbf{0.00} & \textbf{61.20} \\
\midrule
$\times$ & $\checkmark$ & $\checkmark$ & 2.67 & 2.67 & 2.67 & 77.54 & 0.52 & 100.00 & 1.60 & 54.00 \\
\rowcolor{gray!10}
$\checkmark$ & $\times$ & $\checkmark$ & 2.14 & 2.14 & 2.14 & 65.78 & 0.38 & 100.00 & 1.60 & 34.00 \\
$\checkmark$ & $\checkmark$ & $\times$ & 81.82 & 80.74 & 77.00 & 77.54 & 0.00 & 0.00 & 0.00 & 31.20 \\
\rowcolor{gray!10}
$\checkmark$ & $\times$ & $\times$ & 2.14 & 2.14 & 2.14 & 74.87 & 0.00 & 0.00 & 0.00 & 46.00 \\
$\times$ & $\checkmark$ & $\times$ & 3.74 & 3.74 & 3.74 & 75.40 & 0.00 & 0.00 & 0.00 & 60.40 \\
\rowcolor{gray!10}
$\times$ & $\times$ & $\checkmark$ & 4.38 & 4.28 & 4.28 & 73.80 & 1.57 & 100.00 & 6.45 & 40.32 \\
\bottomrule[2pt]
\end{tabular}}
\vspace{-1em}
\end{table}

\textbf{How does each individual term in Eq. \ref{eq:R_it} contribute to DREAM?} As summarized in Table \ref{tab:r_it}, the complete DREAM model delivers the strongest protection, attaining the lowest ASR on EHRAgent, a zero ASR on ReAct, and the highest overall ACC. In comparison, ablating either $p_{ret}$ or $\mathrm{sim}$ produces only moderate degradation on EHRAgent, yet even a small amount of retrieval leakage in ReAct rapidly escalates to an ASR-a of $100.00\%$. This indicates that the retrieval prior and similarity term are particularly critical to preventing subtle leakages from compounding into full failures. In contrast, removing the coherence term $E_{\mathrm{coh}}$ severely undermines EHRAgent, and also leads to a pronounced drop in ReAct ACC.

\begin{wrapfigure}{r}{0.50\textwidth}
\vspace{-1em}
    \centering
    \begin{subfigure}{0.48\linewidth}
        \centering
        \includegraphics[width=\linewidth]{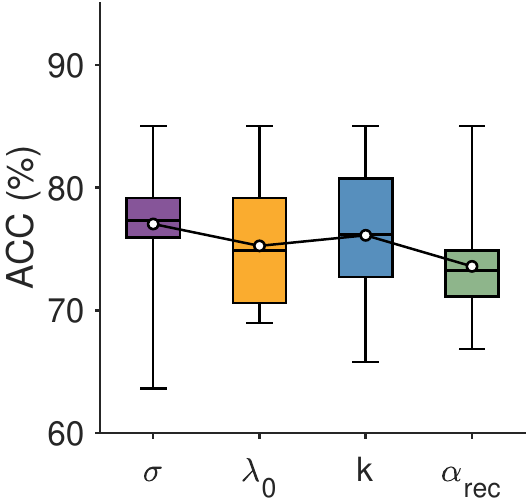}
        \caption{EHRAgent (eICU)}
        \label{fig:ehragent_abl}
    \end{subfigure}\hfill 
    \begin{subfigure}{0.45\linewidth}
        \centering
        \includegraphics[width=\linewidth]{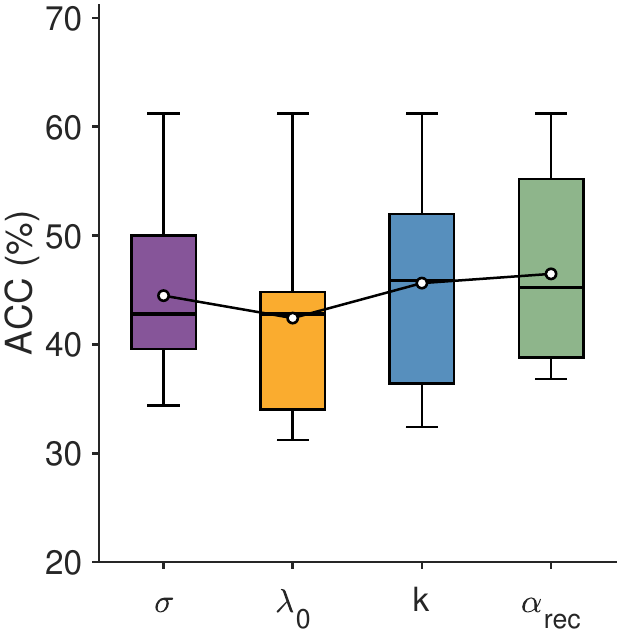}
        \caption{ReAct}
        \label{fig:react_abl}
    \end{subfigure}
    \caption{Performance comparison (ACC) of Ehragent and ReAct under different parameter settings.}
    \label{fig:model_ablation_comparison}
    \vspace{-1em}
\end{wrapfigure}

\textbf{How does DREAM perform under different parameter settings?}
We vary $\sigma$, $\lambda_0$, $k$, and $\alpha_{rec}$ on EHRAgent (eICU) and ReAct with \textit{GPT-5.4-mini}. Full results are reported in Appendix \ref{sec:appendix_sensitivity_analysis} Table~\ref{tab:param_sensitivity_full}. Fig.~\ref{fig:model_ablation_comparison} demonstrates that EHRAgent is comparatively stable once perturbation probing is enabled, with the best ACC $85.03\%$ achieved at $\sigma=0.05$, $\lambda_0=0.4$, $k=10$, and $\alpha_{rec}=0.6$. In contrast, disabling key components degrades performance. For example, setting $\sigma=0$, $k=0$, and $\alpha_{rec}=0$ reduces ACC to $63.64\%$, $65.78\%$, and $67.38\%$, respectively. ReAct is much more sensitive to parameter choice: its best ACC $61.20\%$ is obtained at $\lambda_0=0.2$, $k=15$, and $\alpha_{rec}=0.5$, whereas $\sigma=0$, $\lambda_0=0$, and $\alpha_{rec}=0$ reduce ACC to $36.8\%$, $34.0\%$, and $38.0\%$, and overly large amplification ($k=25$) further lowers ACC to $32.4\%$.

\section{Conclusion}
We presented DREAM, a neuro-inspired, agent-intrinsic memory repair framework for memory-poisoned LLM agents that performs selective functional forgetting without relying on expensive online auditing. By combining perturbation-based probing, multi-dimensional activation diagnosis, and selective forgetting of poison records, DREAM suppresses suspicious memories while preserving benign functionality. Experiments show that DREAM improves robustness across different agent environments and remains efficient compared with baselines. Future work will focus on broader evaluations in more open-ended agent settings.

\bibliographystyle{ieeetr}  
\small
\bibliography{nips}
\normalsize



\newpage
\appendix

\section{Limitations}
\label{apx:limi}
DREAM relies on a multi-dimensional activation probe to distinguish suspicious memories from benign ones. In principle, if an attacker had detailed knowledge of this internal probing mechanism, they could attempt to optimize poisoned memories to remain stable under the probe and thereby weaken the resulting anomaly signal. However, such a probe-aware attack is difficult under the threat models considered in current memory-poisoning work on LLM agents. Existing attacks are typically studied under more constrained capabilities, such as query-only interaction, indirect memory injection through normal user queries, limited write access to the memory store, or black-box access to the agent core, rather than full visibility into the defender's internal monitoring mechanism. Therefore, while probe-aware evasion is a meaningful theoretical limitation, realizing it in practice would require a substantially stronger attacker than those considered in current memory-poisoning settings.

\section{Implementation Details}
\label{apx:ID}
\subsection{Hyperparameters}
The hyperparameters for DREAM and our experiments are reported in Table \ref{tab:hpermeter}.
\begin{table}[htbp]
    \centering
    \setlength{\aboverulesep}{-0.1pt}
\setlength{\belowrulesep}{-0.10pt}
    \caption{Hyperparameter Settings for DREAM}
    \label{tab:hpermeter}
    \begin{tabular}{l|r}
    \toprule[2pt]
    \rowcolor{blue!8}
        \textbf{Parameters} & \textbf{Value} \\
        \midrule[2pt]
         Temperature $\tau$ & 0.05 \\
         The sharpness of coherence decay $\kappa$ & 2.00 \\
         Diagnostic sensitivity $\eta$ & 0.80 \\
         Suspect memory decay rate $\gamma$ & 0.50 \\
         Cold archive threshold $\theta_d$ & 0.10 \\
         Utility Threshold $\theta_c$ & 0.50 \\
         Iterations per forgetting step & 10 \\
         Injection number & 4 \\
         \bottomrule[2pt]
    \end{tabular}
\end{table}

In addition to the above hyperparameters, for EHRAgent we set the noise standard deviation to $\sigma = 0.05$, the base decay rate to $\lambda_0 = 0.40$, the amplification factor to $k = 10$, and the benign recovery rate to $\alpha_{rec} = 0.60$.  For ReAct, we set the noise standard deviation to $\sigma = 0.30$, the base decay rate to $\lambda_0 = 0.20$, the amplification factor to $k = 15$, and the benign recovery rate to $\alpha_{rec} = 0.50$. All experiments are conducted on a server equipped with six NVIDIA A100 GPUs (80 GB memory each).

\subsection{Baselines}
\subsubsection{Distil Classifier}
We adopt a lightweight binary text classifier based on \texttt{DistilBERT-base-uncased} \cite{sanh2019distilbert} as a content-based defense baseline \cite{kumar2023certifying}. The model is fine-tuned to classify memory entries as \emph{safe} or \emph{harmful} according to their textual content alone. Training data is constructed from a composite safe/harmful prompt dataset and split into 80\% training and 20\% validation sets. Inputs are tokenized with \texttt{DistilBertTokenizer} and padded or truncated to 25 tokens. The model is trained for 10 epochs using AdamW ($1\times10^{-5}$, batch size 32), with a \texttt{WeightedRandomSampler} to address class imbalance and gradient clipping at 1.0 for stability. The checkpoint with the lowest validation loss is used for inference.

\subsubsection{Perplexity Filter Baseline}

We adopt a perplexity-based filtering \cite{alon2023detecting} baseline to detect anomalous memory entries according to their textual fluency and coherence. Perplexity scores are computed by a frozen \texttt{LLaMA-3.1-8B-Instruct} model, with all inputs padded or truncated to 100 tokens. For an input sequence $x$, the score is defined as
\begin{equation}
\mathrm{PPL}(x)=\exp\!\left(-\frac{1}{T'}\sum_{t=1}^{T'} \log p(x_t \mid x_{<t})\right),
\end{equation}
where $T'$ is the number of non-padding tokens.

We use a two-stage filtering scheme. First, entries with $\mathrm{PPL}(x)>10000$ are directly discarded as obvious anomalies. Second, for the remaining entries, we apply a dynamic threshold based on the median absolute deviation (MAD): a memory with perplexity $p_i$ is filtered if
\begin{equation}
p_i > m + k\cdot \mathrm{MAD},
\end{equation}
where $m$ is the median perplexity of the batch, $\mathrm{MAD}$ is the median absolute deviation, and $k=1.0$ in our experiments.

\subsubsection{LLM Auditor Baseline}

We adopt an LLM-based auditing baseline that filters the retrieved memory context before it is used by the agent. In our implementation, \texttt{GPT-5.4-mini} and \texttt{Claude-Haiku-4.5} serve as the auditor models. The auditor is instructed to act as a security analyst that removes manipulative, toxic, or logically incoherent content while preserving legitimate information. Given a retrieved memory set $\mathcal{M}_{\mathrm{topK}}$, the auditor returns a cleaned memory context
\begin{equation}
\hat{\mathcal{M}}=\mathrm{Auditor}(\mathcal{M}_{\mathrm{topK}}),
\end{equation}
which is then passed to the agent for downstream reasoning. This baseline represents an online memory-sanitization strategy.

\subsubsection{A-MemGuard}
A-MemGuard \cite{wei2025memguard} is an external online defense framework for memory-augmented LLM agents. It protects the agent through two components: \emph{consensus-based validation}, which detects anomalous memories by comparing parallel reasoning paths induced by different retrieved memories, and a \emph{dual-memory structure}, which stores detected failures as lesson memories to guide future action revision. Given a retrieved memory set $\mathcal{M}_r$, A-MemGuard computes divergence scores
\begin{equation}
d_i = S_{\mathrm{div}}(\rho_i, \mathcal{P}_t),
\end{equation}
where $\rho_i$ is the reasoning path induced by memory $m_i$ and $\mathcal{P}_t$ is the set of all parallel paths. Memories whose paths deviate from the consensus are filtered out, and anomalous paths are further distilled into lesson records for future correction. Thus, A-MemGuard represents a context-aware online auditing baseline with an additional external lesson memory for self-correction.

\section{Results under the REALM retriever.}
\label{apx:realm}
Table~\ref{tab:per_realm} reports the defense performance when the retriever is replaced by REALM. Overall, the main trend remains consistent with the DPR-based results: DREAM is most effective on the structured EHRAgent tasks, while open-ended ReAct remains the more challenging case. Under \textit{GPT-5.4-mini}, DREAM substantially improves robustness on both medical datasets. On eICU, it reduces ASR-r/a/t from $100.00/100.00/99.46$ to $2.14/2.14/1.60$, while also improving ACC from $74.80$ to $78.07$. On MIMIC-III, DREAM similarly lowers ASR-r/a/t to $2.42/1.82/0.00$ and raises ACC from $47.27$ to $54.55$. These results suggest that, even with a different retrieval backbone, DREAM continues to effectively suppress attack propagation in domain-specific agents.

On ReAct (Strategy-QA), the pattern is more nuanced. DREAM still achieves the lowest retrieval-stage and end-to-end attack rates under GPT-5.4-mini, reducing ASR-r from $78.33$ to $0.48$ and ASR-t from $88.78$ to $2.00$. However, ASR-a remains non-negligible ($34.00$), indicating that under open-ended reasoning, residual malicious signals can still propagate inside the reasoning chain even after retrieval has been largely cleaned. This observation is also consistent with the broader baseline behavior in Table~\ref{tab:per_realm}: under REALM, several defenses can partially suppress retrieval-stage attacks on ReAct, but reasoning-stage robustness remains unstable and highly backbone-dependent.

Across methods, PPL and Distil Classifier remain weak on the EHRAgent settings, often leaving attack success near saturation. LLM Auditor and A-MemGuard provide partial mitigation in some cases, but their effectiveness is less consistent across datasets and backbones. Taken together, these results indicate that REALM alone does not eliminate memory poisoning risks; DREAM remains the most reliable defense on structured medical agents and provides the strongest overall suppression of retrieval and end-to-end attacks, while reasoning-stage defense for open-ended agents remains the harder regime.

\begin{table*}[!htbp]
\centering
\caption{Performance evaluation of defense mechanisms against AgentPoison \cite{chen2024agentpoison} attacks across different LLMs and datasets. Lower Attack Success Rate (ASR) in percentage (\%) indicates better performance ($\downarrow$), while higher benign accuracy (ACC) is preferable ($\uparrow$).The retriver is \textbf{REALM}. For ASR, the minimum number in each column is in \textbf{bold}; for ACC, the maximum is in \textbf{bold}.}
\label{tab:per_realm}
\setlength{\aboverulesep}{-0.1pt}
\setlength{\belowrulesep}{-0.10pt}
\renewcommand{\arraystretch}{1.2}
\resizebox{\textwidth}{!}{
\begin{tabular}{ll cccc cccc cccc}
\toprule

\rowcolor{blue!8}
 & & \multicolumn{8}{c}{\textbf{EHRAgent}} & \multicolumn{4}{c}{\textbf{ReAct}} \\
\cmidrule(lr){3-10} \cmidrule(lr){11-14}

\rowcolor{blue!8}
 & & \multicolumn{4}{c}{\textbf{eICU}} & \multicolumn{4}{c}{\textbf{MIMIC-III}} & \multicolumn{4}{c}{\textbf{Strategy-QA}} \\
\cmidrule(lr){3-6} \cmidrule(lr){7-10} \cmidrule(lr){11-14}

\rowcolor{blue!8}
\multirow{-3}{*}{\textbf{LLM}} & \multirow{-3}{*}{\textbf{Defense}} & 
\textbf{ASR-r} & \textbf{ASR-a} & \textbf{ASR-t} & \textbf{ACC} & 
\textbf{ASR-r} & \textbf{ASR-a} & \textbf{ASR-t} & \textbf{ACC} & 
\textbf{ASR-r} & \textbf{ASR-a} & \textbf{ASR-t} & \textbf{ACC} \\
\midrule

\multirow{6}{*}{\textbf{GPT-5.4-mini}} 
& No defense & 100.00 & 100.00 & 99.46 & 74.80 & 100.00 & 92.73 & 23.03 & 47.27 & 78.33 & 93.81 & 88.78 & 34.39 \\
& PPL & 100.00 & 99.47 & 98.93 & 38.50 & 100.00 & 100.00 & 61.21 & 32.12 & 1.96 & 40.00 & \textbf{1.00} & 42.50 \\
& LLM Auditor & 73.26 & 67.91 & 94.12 & 50.80 & 99.39 & 100.00 & 15.15 & 39.39 & 0.95 & 80.00 & 2.00 & 47.00 \\
& Distil Classifier & 100.00 & 99.47 & 98.93 & 45.99 & 100.00 & 0.00 & 65.45 & 38.79 & 2.06 & 37.50 & 1.50 & \textbf{50.50} \\
& A-MemGuard & 64.17 & 72.19 & 53.48 & 44.92 & 72.12 & \textbf{0.00} & 15.38 & 18.27 & 4.44 & 50.00 & 3.08 & 46.55 \\
\rowcolor{gray!15}
\cellcolor{white} & \textbf{DREAM (ours)} & \textbf{2.14} & \textbf{2.14} & \textbf{1.60} & \textbf{78.07} & \textbf{2.42} & 1.82 & \textbf{0.00} & \textbf{54.55} & \textbf{0.48} & \textbf{34.00} & 2.00 & 36.50 \\
\midrule

\multirow{6}{*}{\textbf{Claude-haiku-4.5}} 
& No defense & 100.00 & 86.10 & 59.89 & 42.25 & 100.00 & 72.12 & 13.33 & 39.39 & 94.30 & 72.07 & 60.85 & 23.58 \\
& PPL & 98.93 & 64.17 & 23.53 & 26.20 & 99.39 & 0.61 & 0.00 & 33.33 & 13.41 & 65.24 & 70.83 & 58.50 \\
& LLM Auditor & 100.00 & 89.84 & 47.06 & 32.62 & 61.82 & 0.00 & 7.27 & 36.36 & 9.01 & 60.30 & 6.53 & 62.36 \\
& Distil Classifier & 100.00 & 57.22 & 22.99 & 26.20 & 100.00 & 0.61 & 6.67 & 40.61 & 8.33 & \textbf{60.00} & 1.50 & 61.50 \\
& A-MemGuard & 68.98 & 37.43 & 5.35 & 28.34 & 52.63 & \textbf{0.00} & 0.00 & 36.84 & 8.42 & 67.00 & 1.50 & 62.50 \\
\rowcolor{gray!15}
\cellcolor{white} & \textbf{DREAM (ours)} & \textbf{2.14} & \textbf{6.95} & \textbf{1.07} & \textbf{57.75} & \textbf{2.42} & 3.63 & \textbf{0.00} & \textbf{46.67} & \textbf{4.08} & 75.00 & \textbf{1.50} & \textbf{64.00} \\
\bottomrule

\end{tabular}
}
\end{table*}



\section{Noise distributions for subconscious probing.}
\label{apx:noise}
Given a memory embedding $v_i \in \mathbb{R}^d$, subconscious probing perturbs it as
\begin{equation}
\tilde v_i = v_i + \epsilon_i,
\end{equation}
where $\epsilon_i \in \mathbb{R}^d$ is sampled from a predefined noise distribution. In our ablation study, we consider the following four noise types.

\textbf{Isotropic Gaussian noise.}
The perturbation is sampled independently and identically across all dimensions:
\begin{equation}
\epsilon_i \sim \mathcal{N}(0, \sigma^2 I),
\end{equation}
where $I$ is the identity matrix and $\sigma>0$ controls the perturbation scale. This noise is \emph{isotropic}, meaning that all directions in the embedding space are perturbed equally in expectation. As a result, it provides an unbiased probe of the local geometric stability around each memory embedding.

\textbf{ Uniform noise.}
The perturbation on each dimension is drawn from a bounded uniform distribution:
\begin{equation}
\epsilon_{i,j} \sim \mathbf{U}(-a, a), \qquad j=1,\dots,d,
\end{equation}
where $a>0$ is the noise range. To make it comparable with Gaussian perturbation, $a$ can be chosen such that the variance matches, i.e.,
\begin{equation}
\mathrm{Var}[\epsilon_{i,j}] = \frac{a^2}{3} \approx \sigma^2.
\end{equation}
Unlike Gaussian noise, uniform noise is bounded and does not emphasize rare larger perturbations.

\textbf{Sparse Gaussian noise.}
Only a subset of dimensions is perturbed, while the others remain unchanged:
\begin{equation}
\epsilon_i = m_i \odot z_i,
\end{equation}
where $\odot$ denotes element-wise multiplication, $z_i \sim \mathcal{N}(0,\sigma^2 I)$, and the mask $m_i \in \{0,1\}^d$ is sampled as
\begin{equation}
m_{i,j} \sim \mathrm{Bernoulli}(p).
\end{equation}
Here $p \in (0,1]$ is the activation probability of each dimension. Sparse Gaussian noise probes only localized coordinates of the embedding, producing a more selective but less globally informative perturbation.

\textbf{Anisotropic Gaussian noise.}
The perturbation follows a Gaussian distribution with direction-dependent covariance:
\begin{equation}
\epsilon_i \sim \mathcal{N}(0, \Sigma),
\end{equation}
where $\Sigma \in \mathbb{R}^{d \times d}$ is a positive semi-definite covariance matrix. A common special case is diagonal anisotropy,
\begin{equation}
\Sigma = \mathrm{diag}(\sigma_1^2,\sigma_2^2,\dots,\sigma_d^2),
\end{equation}
which assigns different perturbation magnitudes to different dimensions. Compared with isotropic Gaussian noise, anisotropic Gaussian noise emphasizes certain directions in the embedding space and may therefore introduce directional bias in stability probing. 
For a fair comparison, different noise types are calibrated to have approximately matched expected perturbation energy, i.e.,
\begin{equation}
\mathbb{E}\|\epsilon_i\|_2^2 \approx d \sigma^2.
\end{equation}

\paragraph{Interpretation.}
These four perturbations differ in how they explore the local neighborhood around a memory embedding. Isotropic Gaussian noise perturbs all directions evenly and thus provides the most geometry-preserving probe of local stability. Uniform noise is similarly dense but bounded. Sparse Gaussian noise perturbs only partial coordinates and may miss global fragility patterns. Anisotropic Gaussian noise is dense but directionally biased, which can overemphasize certain subspaces. This difference helps explain why isotropic Gaussian noise yields the strongest benign utility in our ablation while maintaining consistently low attack success.

\begin{table*}[htbp]
\centering
\caption{Performance of different noise type.}
\label{tab:noise_types}
\setlength{\aboverulesep}{-0.1pt}
\setlength{\belowrulesep}{-0.1pt}
\renewcommand{\arraystretch}{1.2}
\resizebox{0.7\textwidth}{!}{
\begin{tabular}{l cccc}
\toprule
\rowcolor{blue!8}
Noise Type & \textbf{ASR-r} & \textbf{ASR-a} & \textbf{ASR-t} & \textbf{ACC} \\

Isotropic Gussian & 2.14 & 2.14 & 2.14 & \textbf{85.03} \\
\rowcolor{gray!10}
Uniform & 2.14 & 2.14 & 2.14 & 77.01 \\
Sparse Gaussian & 2.14 & 2.67 & 2.14 & 77.54 \\
\rowcolor{gray!10}
Anisotropic Gussian & 2.14 & 2.14 & 2.14 & 81.28 \\
\bottomrule
\end{tabular}
}
\end{table*}

As shown in Table~\ref{tab:noise_types}, all tested noise types keep the attack success rates at a comparably low level, indicating that the overall defense mechanism is not overly sensitive to the specific perturbation family. However, their impact on benign performance differs more clearly. Isotropic Gaussian noise achieves the best ACC ($85.03\%$), outperforming Uniform noise ($77.01\%$), Sparse Gaussian noise ($77.54\%$), and Anisotropic Gaussian noise ($81.28\%$). In particular, Sparse Gaussian noise also slightly increases ASR-a to $2.67\%$, suggesting that overly localized perturbations provide a less reliable probing signal. These results support our choice of isotropic Gaussian noise, which offers the best balance between preserving benign utility and maintaining consistently low attack success.

\section{Analysis for \texorpdfstring{$R_i(t)$}{R\_i(t)}}
\label{apx:r_it}

\begin{figure}[!htbp]
    \centering
    \includegraphics[width=\linewidth]{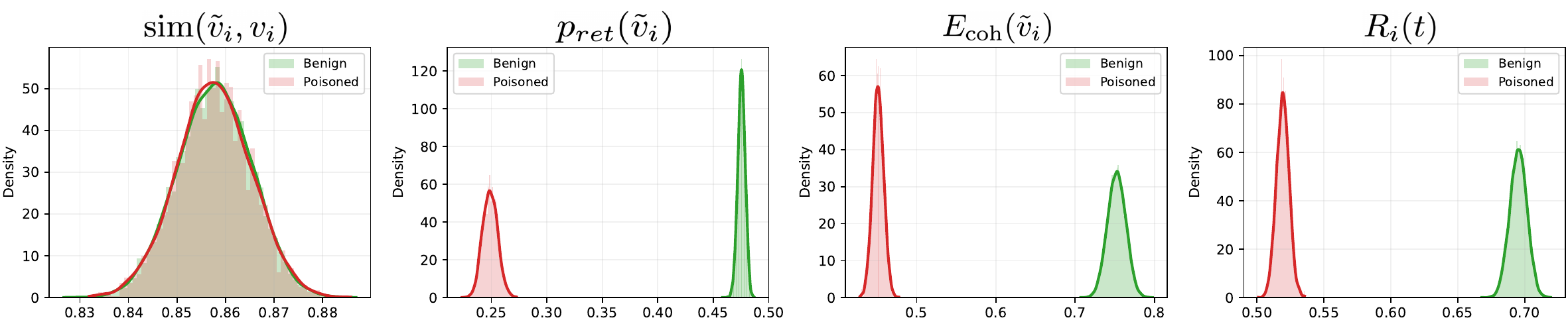}
    \caption{Empirical distributions of the three component signals used in $R_i(t)$. While $\mathrm{sim}(\tilde{v}_i, v_i)$ shows substantial overlap between benign and poisoned memories, $p_{ret}$ and $E_{coh}$ exhibit clearer shifts. This heterogeneous behavior motivates their aggregation into the multi-dimensional activation intensity $R_i(t)$.}
    \label{fig:single_component_distributions}
\end{figure}

As shown in Fig.~\ref{fig:single_component_distributions}, the three components exhibit clearly different discriminative behaviors. The $\mathrm{sim}(\tilde{v}_i, v_i)$ distribution of benign and poisoned memories remains heavily overlapped, indicating that local geometric stability alone is insufficient for reliable diagnosis. In contrast, $p_{ret}$ and $E_{coh}$ show much stronger distributional shifts, suggesting that poisoned memories are less competitive in retrieval and weaker in semantic coherence under perturbation. Importantly, however, these two components capture different aspects of abnormality: $p_{ret}$ reflects global retrieval competitiveness, whereas $E_{coh}$ reflects semantic consistency. As further shown in Fig.~\ref{fig:single_component_distributions}, after aggregating these heterogeneous cues into the multi-dimensional activation intensity $R_i(t)$, the separation between benign and poisoned memories becomes substantially clearer than in any individual weak component such as $\mathrm{sim}(\tilde{v}_i, v_i)$. Therefore, as also demonstrated in Table \ref{tab:r_it}, the key observation is not that any single component is universally sufficient, but that the three components provide heterogeneous yet complementary evidence. This precisely justifies the definition of $R_i(t)$ as a unified and more robust diagnostic score.




\section{Forgetting dynamics of \texorpdfstring{$R_{i}$ and $w_{i}$}{R\_i and w\_i}.}
\label{apx:wd}
\begin{figure*}[t]
    \centering
    \includegraphics[width=0.82\textwidth]{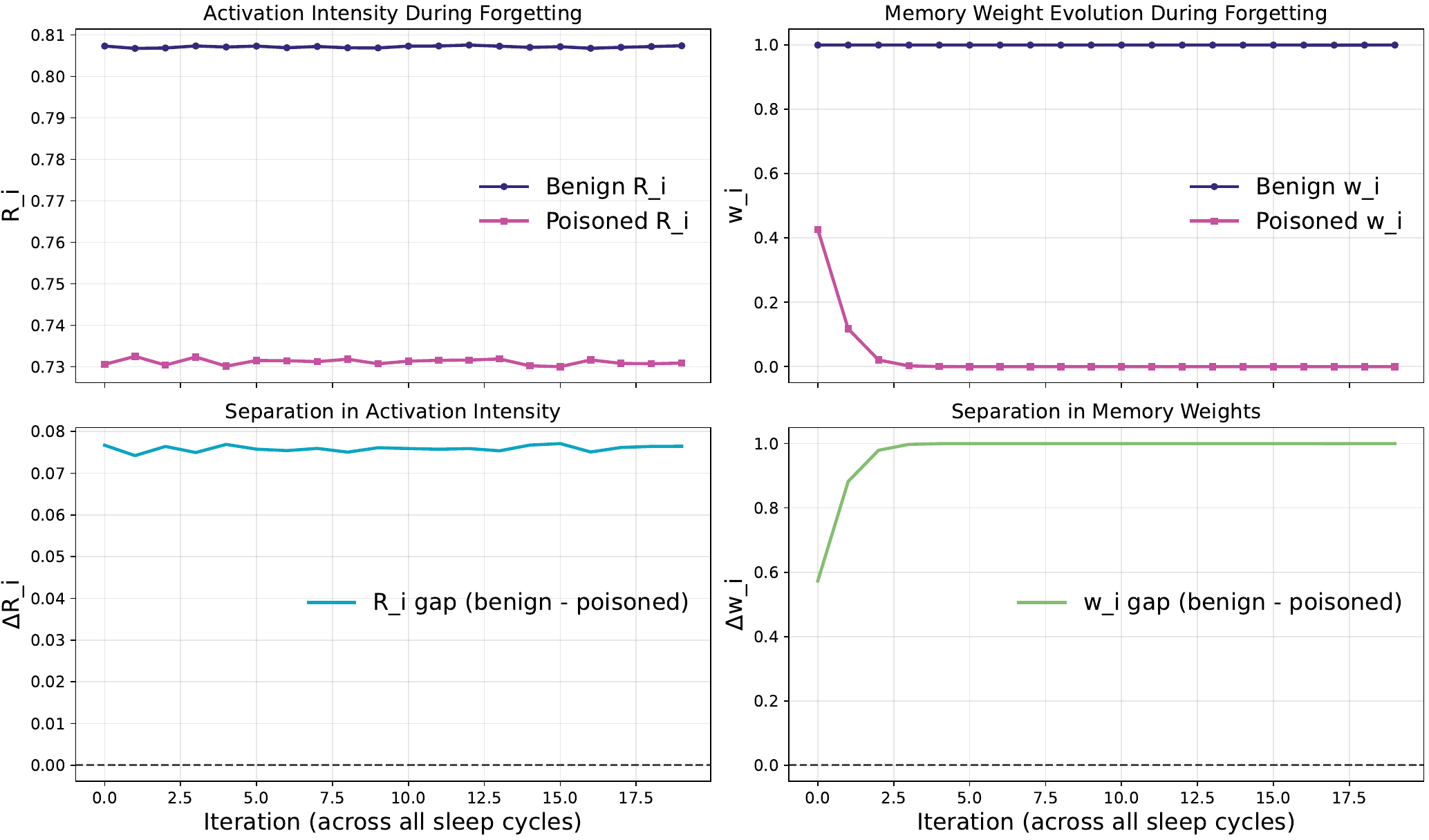}
    \caption{Memory forgetting dynamics of benign and poisoned memories across repeated sleep cycles. The top row shows the evolution of activation intensity $R_{i}(t)$ and memory weight $w_{i}(t)$, respectively. The bottom row reports the corresponding separation gaps between benign and poisoned memories. While the $R_{i}(t)$ gap remains consistently positive, the $w_{i}(t)$ gap rapidly expands and saturates, indicating that stable activation-level differences are progressively translated into stronger functional separation during forgetting.}
    \label{fig:appendix_forgetting_dynamics}
\end{figure*}

Fig.~\ref{fig:appendix_forgetting_dynamics} visualizes the evolution of activation intensity and memory weights during repeated sleep cycles. As shown in the top-left panel, benign memories consistently maintain higher $R_{i}(t)$ values than poisoned ones, and this gap remains stably positive throughout the entire forgetting process. This indicates that the proposed multi-dimensional activation intensity provides a persistent diagnostic signal rather than a transient fluctuation. In turn, the top-right panel shows that this stable discrepancy in $R_{i}(t)$ is translated into sharply different weight trajectories: benign memories remain close to unit weight, whereas poisoned memories are rapidly attenuated toward zero within only a few iterations. The bottom panels further highlight this effect quantitatively. While the $R_{i}(t)$ gap stays relatively constant across cycles, the $w_{i}(t)$ gap quickly increases and saturates near its maximum, showing that DREAM converts a stable activation-level separation into an increasingly pronounced functional separation in memory influence. Together, these results illustrate the intended behavior of selective functional forgetting: suspicious memories are progressively suppressed, whereas benign memories are stably preserved.

\section{Online Consolidation Signal and Memory-level Utility Attribution}
\label{apx:online_consolidation}
\textbf{Online Consolidation Signal.}
Let $c_i(t)\in[0,1]$ denote the online consolidation signal of memory $v_i$, which reflects its historical cross-task utility during the agent's online phase. Intuitively, $c_i(t)$ serves as a long-term trust indicator: memories that are repeatedly retrieved and associated with successful task completion should be less aggressively attenuated during subsequent repair cycles.

To avoid overreacting to transient task-specific fluctuations, we update the consolidation signal using an exponential moving average. Specifically, for online cycle $t$ and query $q_t$, let $\mathcal{M}_{\mathrm{topK}}(q_t)$ denote the retrieved memory set. We update $c_i(t)$ only when $v_i$ is actually retrieved:
\begin{equation}
c_i(t)=
\begin{cases}
\delta \, c_i(t-1) + (1-\delta)\, u_i(t), & v_i \in \mathcal{M}_{\mathrm{topK}}(q_t),\\
c_i(t-1), & \text{otherwise},
\end{cases}
\label{eq:ci_appendix}
\end{equation}
where $\delta\in(0,1)$ is a momentum coefficient controlling temporal smoothness.

Here, $u_i(t)\in[0,1]$ denotes the memory-level utility assigned to $v_i$ at online cycle $t$. Since multiple retrieved memories may jointly contribute to the final task outcome, we adopt a lightweight attribution scheme that decomposes task-level utility into memory-level contributions. Let $U_t\in[0,1]$ denote the task-level utility signal achieved at cycle $t$ (i.e., task success signal). We define
\begin{equation}
u_i(t)=a_i(t)\cdot U_t,
\label{eq:ui_appendix}
\end{equation}
where $a_i(t)$ is a normalized attribution weight over the retrieved set:
\begin{equation}
a_i(t)=
\begin{cases}
\dfrac{\exp(\mathrm{Score}(q_t,v_i)/\tau_u)}
{\sum_{v_j\in \mathcal{M}_{\mathrm{topK}}(q_t)} \exp(\mathrm{Score}(q_t,v_j)/\tau_u)},
& v_i\in \mathcal{M}_{\mathrm{topK}}(q_t),\\[1.2em]
0, & \text{otherwise}.
\end{cases}
\label{eq:ai_appendix}
\end{equation}
Here, $\mathrm{Score}(q_t,v_i)$ denotes the retrieval score of memory $v_i$ under query $q_t$, and $\tau_u>0$ is a temperature parameter. This formulation assigns larger utility to memories that are both retrieved and more strongly implicated in successful task completion, while avoiding the computational overhead of leave-one-out re-evaluation.

Overall, the online consolidation signal $c_i(t)$ provides a complementary utility-aware factor to the offline anomaly diagnosis: suspicious memories with persistently low utility remain vulnerable to attenuation, whereas benign but atypical memories can recover long-term trust through repeated successful use.

\section{Detailed Sensitivity Analysis on Defense Parameters}
\label{sec:appendix_sensitivity_analysis}

\begin{table*}[htbp]
\centering
\caption{Parameter sensitivity analysis of DREAM on EHRAgent (eICU) and ReAct. All metrics are reported in percentages (\%). The lower ASR is better ($\downarrow$), while the higher ACC is better ($\uparrow$). }
\label{tab:param_sensitivity_full}
\setlength{\aboverulesep}{0pt}
\setlength{\belowrulesep}{0pt}
\renewcommand{\arraystretch}{1.15}
\resizebox{\textwidth}{!}{
\begin{tabular}{cc cccc cccc}
\toprule
\rowcolor{blue!8}
\textbf{Parameter} & \textbf{Value}
& \multicolumn{4}{c}{\textbf{EHRAgent (eICU)}}
& \multicolumn{4}{c}{\textbf{ReAct}} \\
\cmidrule(lr){3-6} \cmidrule(lr){7-10}
\rowcolor{blue!8}
& & \textbf{ASR-r} & \textbf{ASR-a} & \textbf{ASR-t} & \textbf{ACC}
& \textbf{ASR-r} & \textbf{ASR-a} & \textbf{ASR-t} & \textbf{ACC} \\
\midrule

\multirow{11}{*}{$\sigma$}
& 0.00 & 99.47 & 99.47 & 99.47 & 63.64 & 0.42 & 100.00 & 1.60 & 36.80 \\
& 0.05 & 2.14  & 2.14  & 2.14  & \textbf{85.03} & 0.45 & 100.00 & 1.60 & 42.80 \\
& 0.10 & 2.14  & 2.14  & 2.14  & 75.94 & 0.32 & 100.00 & 1.20 & 50.00 \\
& 0.15 & 2.14  & 2.14  & 2.14  & 77.01 & 0.91 & 80.00  & 1.60 & 34.40 \\
& 0.20 & 2.14  & 2.14  & 2.14  & 81.82 & 0.40 & 100.00 & 1.60 & 39.60 \\
& 0.25 & 2.14  & 2.14  & 2.14  & 79.14 & 0.42 & 100.00 & 1.60 & 46.40 \\
& 0.30 & 2.14  & 2.14  & 2.14  & 77.54 & 0.00 & 0.00   & 0.00 & \textbf{61.20} \\
& 0.35 & 2.14  & 2.14  & 2.14  & 77.33 & 0.00 & 0.00   & 20.40 & 42.80 \\
& 0.40 & 2.14  & 2.14  & 2.14  & 77.01 & 0.32 & 100.00 & 1.20 & 41.29 \\
& 0.45 & 2.14  & 2.14  & 2.14  & 75.40 & 0.36 & 100.00 & 1.60 & 40.80 \\
& 0.50 & 2.14  & 2.14  & 2.14  & 77.54 & 0.00 & 0.00   & 0.00 & 53.20 \\
\midrule

\multirow{11}{*}{$\lambda_0$}
& 0.00  & 100.00 & 100.00 & 100.00 & 74.87 & 0.00 & 0.00   & 0.00 & 34.00 \\
& 0.10 & 2.14   & 2.14   & 2.14   & 79.14 & 0.40 & 100.00 & 1.60 & 44.80 \\
& 0.20 & 2.14   & 2.14   & 2.14   & 68.98 & 0.00 & 0.00   & 0.00 & \textbf{61.20} \\
& 0.30 & 3.74   & 3.74   & 3.74   & 74.33 & 0.41 & 100.00 & 1.60 & 42.80 \\
& 0.40 & 2.14   & 2.14   & 2.14   & \textbf{85.03} & 0.00 & 0.00   & 0.00 & 38.80 \\
& 0.50 & 2.14   & 2.14   & 2.14   & 76.47 & 0.43 & 100.00 & 1.60 & 44.80 \\
& 0.60 & 2.14   & 2.14   & 2.14   & 79.14 & 0.36 & 100.00 & 1.60 & 31.20 \\
& 0.70 & 100.00 & 98.93  & 98.93  & 72.19 & 0.39 & 100.00 & 0.40 & 34.00 \\
& 0.80 & 100.00 & 98.93  & 98.93  & 77.54 & 0.00 & 0.00   & 0.00 & 36.61 \\
& 0.90 & 2.14   & 0.534  & 0.534  & 69.52 & 0.00 & 0.00   & 0.00 & 54.40 \\
& 1.00 & 2.14   & 2.14   & 2.14   & 70.59 & 0.42 & 100.00 & 1.60 & 44.00 \\
\midrule

\multirow{6}{*}{$k$}
& 0  & 2.14 & 2.14 & 2.14 & 65.78 & 0.00 & 0.00   & 0.00 & 52.00 \\
& 5  & 2.14 & 2.14 & 2.14 & 80.74 & 0.43 & 100.00 & 1.60 & 44.80 \\
& 10 & 2.14 & 2.14 & 2.14 & \textbf{85.03} & 0.00 & 0.00   & 0.00 & 46.99 \\
& 15 & 2.14 & 2.14 & 2.14 & 78.07 & 0.00 & 0.00   & 0.00 & \textbf{61.20} \\
& 20 & 2.14 & 2.14 & 2.14 & 74.33 & 0.42 & 100.00 & 1.60 & 36.40 \\
& 25 & 2.14 & 2.14 & 2.14 & 72.74 & 1.48 & 28.57  & 0.80 & 32.40 \\
\midrule

\multirow{10}{*}{$\alpha_{rec}$}
& 0.00   & 2.14 & 2.14 & 2.14 & 67.38 & 0.37 & 100.00 & 1.60 & 38.00 \\
& 0.10 & 2.14 & 2.14 & 2.14 & 77.01 & 0.48 & 100.00 & 2.00 & 46.00 \\
& 0.20 & 2.14 & 2.14 & 2.14 & 74.87 & 0.00 & 0.00   & 0.00 & 55.20 \\
& 0.30 & 2.14 & 2.14 & 2.14 & 73.80 & 0.00 & 0.00   & 0.00 & 41.20 \\
& 0.40 & 2.14 & 2.14 & 2.14 & 73.26 & 0.00 & 0.00   & 0.00 & 36.80 \\
& 0.50 & 2.14 & 2.14 & 2.14 & 73.28 & 0.00 & 0.00   & 0.00 & \textbf{61.20} \\
& 0.60 & 2.14 & 2.14 & 2.14 & \textbf{85.03} & 0.52 & 100.00 & 1.60 & 56.80 \\
& 0.70 & 2.14 & 2.14 & 2.14 & 73.26 & 0.71 & 40.00  & 0.80 & 38.80 \\
& 0.80 & 2.14 & 2.14 & 2.14 & 71.12 & 0.00 & 0.00   & 0.00 & 46.40 \\
& 0.90 & 2.14 & 2.14 & 2.14 & 66.84 & 0.42 & 100.00 & 1.60 & 44.40 \\
\bottomrule
\end{tabular}
}
\end{table*}

To provide a more complete view of DREAM's internal behavior, Table~\ref{tab:param_sensitivity_full} reports the full parameter sensitivity results for $\sigma$, $\lambda_0$, $k$, and $\alpha_{rec}$ on both EHRAgent and ReAct.

\paragraph{Effect of perturbation scale ($\sigma$).}
The perturbation scale controls the strength of subconscious probing. On EHRAgent, setting $\sigma=0$ causes a near-complete defense collapse (ASR-r/a/t $=99.47/99.47/99.47$), showing that perturbation probing is indispensable. Once probing is enabled, the attack rates remain uniformly low (all $2.14\%$), while ACC varies within a narrower range and peaks at $85.03\%$ when $\sigma=0.05$. ReAct is substantially more sensitive to $\sigma$: small perturbations often fail to suppress reasoning-stage attacks, whereas intermediate values such as $\sigma=0.3$ and $\sigma=0.5$ reduce all ASRs to zero and achieve the best ACC values of $61.20\%$ and $53.20\%$, respectively. This suggests that probing is necessary for both agents, but its effective scale is much more delicate for open-ended reasoning.

\paragraph{Effect of base decay rate ($\lambda_0$).}
The parameter $\lambda_0$ sets the baseline attenuation strength. On EHRAgent, moderate values ($0.1$--$0.6$) keep ASRs low, with the best ACC of $85.03\%$ achieved at $\lambda_0=0.4$. However, overly weak or poorly calibrated decay can still be harmful: at $\lambda_0=0$, defense collapses completely, while at $\lambda_0=0.7$ and $0.8$ the attack rates rise again to near-saturation. On ReAct, $\lambda_0$ is the most sensitive parameter. Several values (e.g., $0.2$, $0.4$, $0.8$, and $0.9$) suppress all ASRs to zero, but the corresponding ACC varies widely, from $36.61\%$ to $61.20\%$. The best configuration is $\lambda_0=0.2$, indicating that decay calibration is critical for preserving utility in open-ended agents.

\paragraph{Effect of amplification factor ($k$).}
The factor $k$ determines how strongly suspect scores amplify attenuation. EHRAgent is relatively stable in ASR across all tested $k$, but ACC is clearly optimized by moderate amplification, peaking at $85.03\%$ when $k=10$. Setting $k=0$ lowers ACC to $65.78\%$, showing that amplification contributes meaningfully to utility-preserving filtering. ReAct again exhibits a more delicate trade-off: moderate settings such as $k=10$ and $15$ drive all ASRs to zero, with the best ACC of $61.20\%$ obtained at $k=15$, whereas an overly large amplification ($k=25$) sharply degrades ACC to $32.40\%$ and also reintroduces nonzero attack success. This confirms that aggressive amplification can over-prune useful reasoning context in open-ended agents.

\paragraph{Effect of recovery rate ($\alpha_{rec}$).}
The recovery rate controls the reconsolidation of benign but temporarily attenuated memories. On EHRAgent, ASRs remain unchanged across $\alpha_{rec}$, but ACC is strongly affected: disabling recovery ($\alpha_{rec}=0$) reduces ACC to $67.38\%$, while the best performance is achieved at $\alpha_{rec}=0.6$ with $85.03\%$. On ReAct, recovery is even more consequential. Several intermediate settings ($\alpha_{rec}=0.2$, $0.3$, $0.4$, $0.5$, and $0.8$) suppress all ASRs to zero, but ACC varies substantially, peaking at $61.20\%$ when $\alpha_{rec}=0.5$. When recovery is removed ($\alpha_{rec}=0$), ACC drops to $38.00\%$ and ASR-a returns to $100\%$. These results indicate that reconsolidation is not merely auxiliary, but essential for restoring benign functionality after attenuation, especially in agents with more fragile reasoning trajectories.

\section{Use of Large Language Models (LLMs)}
Large language models were used exclusively for linguistic polishing, such as improving grammar and readability. They were not involved in the formulation of research ideas, study design, implementation, data analysis, or reference selection. All scientific contributions, technical claims, and interpretations in this paper were conceived, authored, and rigorously validated by the human authors.

\newpage
\section*{NeurIPS Paper Checklist}
\begin{enumerate}

\item {\bf Claims}
    \item[] Question: Do the main claims made in the abstract and introduction accurately reflect the paper's contributions and scope?
    \item[] Answer: \answerYes{} 
    \item[] Justification: In the abstract and introduction, we have thoroughly detailed the background, motivation, scope, main experimental results, and contributions of our work.

\item {\bf Limitations}
    \item[] Question: Does the paper discuss the limitations of the work performed by the authors?
    \item[] Answer: \answerYes{} 
    \item[] Justification: We provide discussions the limitation of our method in Appendix \ref{apx:limi}.

\item {\bf Theory assumptions and proofs}
    \item[] Question: For each theoretical result, does the paper provide the full set of assumptions and a complete (and correct) proof?
    \item[] Answer: \answerYes{} 
    \item[] Justification: The assumptions in this paper are properly referenced.

    \item {\bf Experimental result reproducibility}
    \item[] Question: Does the paper fully disclose all the information needed to reproduce the main experimental results of the paper to the extent that it affects the main claims and/or conclusions of the paper (regardless of whether the code and data are provided or not)?
    \item[] Answer: \answerYes{}
    \item[] Justification: The paper has disclosed all the information in the method, experiment section and appendix. The data and code are also publicly accessible and provided with detailed procedures to reproduce the results reported in this paper.

\item {\bf Open access to data and code}
    \item[] Question: Does the paper provide open access to the data and code, with sufficient instructions to faithfully reproduce the main experimental results, as described in supplemental material?
    \item[] Answer: \answerYes{} 
    \item[] Justification: The code and data are publicly accessible.

\item {\bf Experimental setting/details}
    \item[] Question: Does the paper specify all the training and test details (e.g., data splits, hyperparameters, how they were chosen, type of optimizer) necessary to understand the results?
    \item[] Answer: \answerYes{} 
    \item[] Justification: Details of both training and test are disclosed.

\item {\bf Experiment statistical significance}
    \item[] Question: Does the paper report error bars suitably and correctly defined or other appropriate information about the statistical significance of the experiments?
    \item[] Answer: \answerYes{} 
    \item[] Justification: \justificationTODO{}
    \item[] Guidelines: We adopt standard evaluation datasets, LLM agents, and metrics, which are accompanied by statistical significance.
    \begin{itemize}
        \item The answer \answerNA{} means that the paper does not include experiments.
        \item The authors should answer \answerYes{} if the results are accompanied by error bars, confidence intervals, or statistical significance tests, at least for the experiments that support the main claims of the paper.
        \item The factors of variability that the error bars are capturing should be clearly stated (for example, train/test split, initialization, random drawing of some parameter, or overall run with given experimental conditions).
        \item The method for calculating the error bars should be explained (closed form formula, call to a library function, bootstrap, etc.)
        \item The assumptions made should be given (e.g., Normally distributed errors).
        \item It should be clear whether the error bar is the standard deviation or the standard error of the mean.
        \item It is OK to report 1-sigma error bars, but one should state it. The authors should preferably report a 2-sigma error bar than state that they have a 96\% CI, if the hypothesis of Normality of errors is not verified.
        \item For asymmetric distributions, the authors should be careful not to show in tables or figures symmetric error bars that would yield results that are out of range (e.g., negative error rates).
        \item If error bars are reported in tables or plots, the authors should explain in the text how they were calculated and reference the corresponding figures or tables in the text.
    \end{itemize}

\item {\bf Experiments compute resources}
    \item[] Question: For each experiment, does the paper provide sufficient information on the computer resources (type of compute workers, memory, time of execution) needed to reproduce the experiments?
    \item[] Answer: \answerYes{} 
    \item[] Justification: The experiments are sufficiently discussed to be run by others.
    
\item {\bf Code of ethics}
    \item[] Question: Does the research conducted in the paper conform, in every respect, with the NeurIPS Code of Ethics \url{https://neurips.cc/public/EthicsGuidelines}?
    \item[] Answer: \answerYes{} 
    \item[] Justification: The conducted research conforms in every respect with the NeurIPS Code of Ethics.

\item {\bf Broader impacts}
    \item[] Question: Does the paper discuss both potential positive societal impacts and negative societal impacts of the work performed?
    \item[] Answer: \answerYes{} 
    \item[] Justification: Both potential positive societal impacts and negative societal impacts of the work have been discussed.
    
\item {\bf Safeguards}
    \item[] Question: Does the paper describe safeguards that have been put in place for responsible release of data or models that have a high risk for misuse (e.g., pre-trained language models, image generators, or scraped datasets)?
    \item[] Answer: \answerYes{} 
    \item[] Justification: The paper has discussed safeguards for responsible release of data or models that have a high risk for misuse.

\item {\bf Licenses for existing assets}
    \item[] Question: Are the creators or original owners of assets (e.g., code, data, models), used in the paper, properly credited and are the license and terms of use explicitly mentioned and properly respected?
    \item[] Answer: \answerYes{} 
    \item[] Justification: All the mentioned previous work are properly cited.

\item {\bf New assets}
    \item[] Question: Are new assets introduced in the paper well documented and is the documentation provided alongside the assets?
    \item[] Answer: \answerNA{} 
    \item[] Justification: The code will be made public upon acceptance.

\item {\bf Crowdsourcing and research with human subjects}
    \item[] Question: For crowdsourcing experiments and research with human subjects, does the paper include the full text of instructions given to participants and screenshots, if applicable, as well as details about compensation (if any)? 
    \item[] Answer: \answerNA{} 
    \item[] Justification: The paper does not involve crowdsourcing nor research with human subjects.

\item {\bf Institutional review board (IRB) approvals or equivalent for research with human subjects}
    \item[] Question: Does the paper describe potential risks incurred by study participants, whether such risks were disclosed to the subjects, and whether Institutional Review Board (IRB) approvals (or an equivalent approval/review based on the requirements of your country or institution) were obtained?
    \item[] Answer: \answerNA{} 
    \item[] Justification: The paper does not involve crowdsourcing nor research with human subjects.

\item {\bf Declaration of LLM usage}
    \item[] Question: Does the paper describe the usage of LLMs if it is an important, original, or non-standard component of the core methods in this research? Note that if the LLM is used only for writing, editing, or formatting purposes and does \emph{not} impact the core methodology, scientific rigor, or originality of the research, declaration is not required.
    \item[] Answer: \answerNA{} 
    \item[] Justification: This research does not involve LLMs as any important, original, or non-standard components.

\end{enumerate}

\end{document}